\documentclass[sigconf]{acmart}

\copyrightyear{2023}
\acmYear{2023}
\setcopyright{acmlicensed}\acmConference[WSDM '23]{Proceedings of the Sixteenth ACM International Conference on Web Search and Data Mining}{February 27-March 3, 2023}{Singapore, Singapore}
\acmBooktitle{Proceedings of the Sixteenth ACM International Conference on Web Search and Data Mining (WSDM '23), February 27-March 3, 2023, Singapore, Singapore}
\acmPrice{15.00}
\acmDOI{10.1145/3539597.3570429}
\acmISBN{978-1-4503-9407-9/23/02}

\AtBeginDocument{%
  }

\usepackage{booktabs}
\usepackage{multirow}
\usepackage{makecell}
\usepackage[normalem]{ulem}
\useunder{\uline}{\ul}{}
\usepackage[ruled,linesnumbered]{algorithm2e}
\usepackage{subfigure} 
\usepackage{enumitem} 
\usepackage{amsmath, amsbsy}

\begin{document}

\begin{sloppypar}

\title{Position-Aware Subgraph Neural Networks with Data-Efficient Learning}

\author{Chang Liu}
\email{isonomialiu@sjtu.edu.cn}
\orcid{0000-0001-9341-6002}
\affiliation{%
  \institution{Department of Computer Science and Engineering}
    \institution{Shanghai Jiao Tong University}  
  \city{Shanghai}
  \country{China}
}

\author{Yuwen Yang}
\email{youngfish@sjtu.edu.cn}
\affiliation{%
  \institution{Department of Computer Science and Engineering}
    \institution{Shanghai Jiao Tong University}  
  \city{Shanghai}
  \country{China}
}

\author{Zhe Xie}
\email{xiezhe20001128@sjtu.edu.cn}
\affiliation{%
  \institution{Department of Computer Science and Engineering}
    \institution{Shanghai Jiao Tong University}  
  \city{Shanghai}
  \country{China}
}

\author{Hongtao Lu}
\authornotemark[1]
\email{htlu@sjtu.edu.cn}
\affiliation{%
  \institution{Department of Computer Science and Engineering}
    \institution{Shanghai Jiao Tong University}  
  \city{Shanghai}
  \country{China}
}

\author{Yue Ding}
\email{dingyue@sjtu.edu.cn}
\authornote{Corresponding author}
\affiliation{%
  \institution{Department of Computer Science and Engineering}
    \institution{Shanghai Jiao Tong University}  
  \city{Shanghai}
  \country{China}
}

\begin{abstract}
Data-efficient learning on graphs (GEL) is essential in real-world applications. Existing GEL methods focus on learning useful representations for nodes, edges, or entire graphs with ``small'' labeled data. But the problem of data-efficient learning for subgraph prediction has not been explored. The challenges of this problem lie in the following aspects: 
1) It is crucial for subgraphs to learn positional features to acquire structural information in the base graph in which they exist. Although the existing subgraph neural network method is capable of learning disentangled position encodings, the overall computational complexity is very high.
2) Prevailing graph augmentation methods for GEL, including rule-based, sample-based, adaptive, and automated methods, are not suitable for augmenting subgraphs because a subgraph contains fewer nodes but richer information such as position, neighbor, and structure. Subgraph augmentation is more susceptible to undesirable perturbations. 
3) Only a small number of nodes in the base graph are contained in subgraphs, which leads to a potential ``bias'' problem that the subgraph representation learning is dominated by these ``hot'' nodes. By contrast, the remaining nodes fail to be fully learned, which reduces the generalization ability of subgraph representation learning.
In this paper, we aim to address the challenges above and propose a Position-Aware Data-Efficient Learning framework for subgraph neural networks called PADEL. 
Specifically, we propose a novel node position encoding method that is anchor-free, and design a new generative subgraph augmentation method based on a diffused variational subgraph autoencoder, and we propose exploratory and exploitable views for subgraph contrastive learning. 
Extensive experiment results on three real-world datasets show the superiority of our proposed method over state-of-the-art baselines.
\end{abstract}

\ccsdesc[500]{Computing methodologies~Supervised learning by classification}

\keywords{subgraph neural networks, contrastive learning, generative model, data-efficient learning}

\maketitle

\section{Introduction}
\label{sec:intro}

\begin{figure}[h] 
\setlength{\abovecaptionskip}{0cm}
\setlength{\belowcaptionskip}{0cm}
\centering 
\includegraphics[width=2.8in]{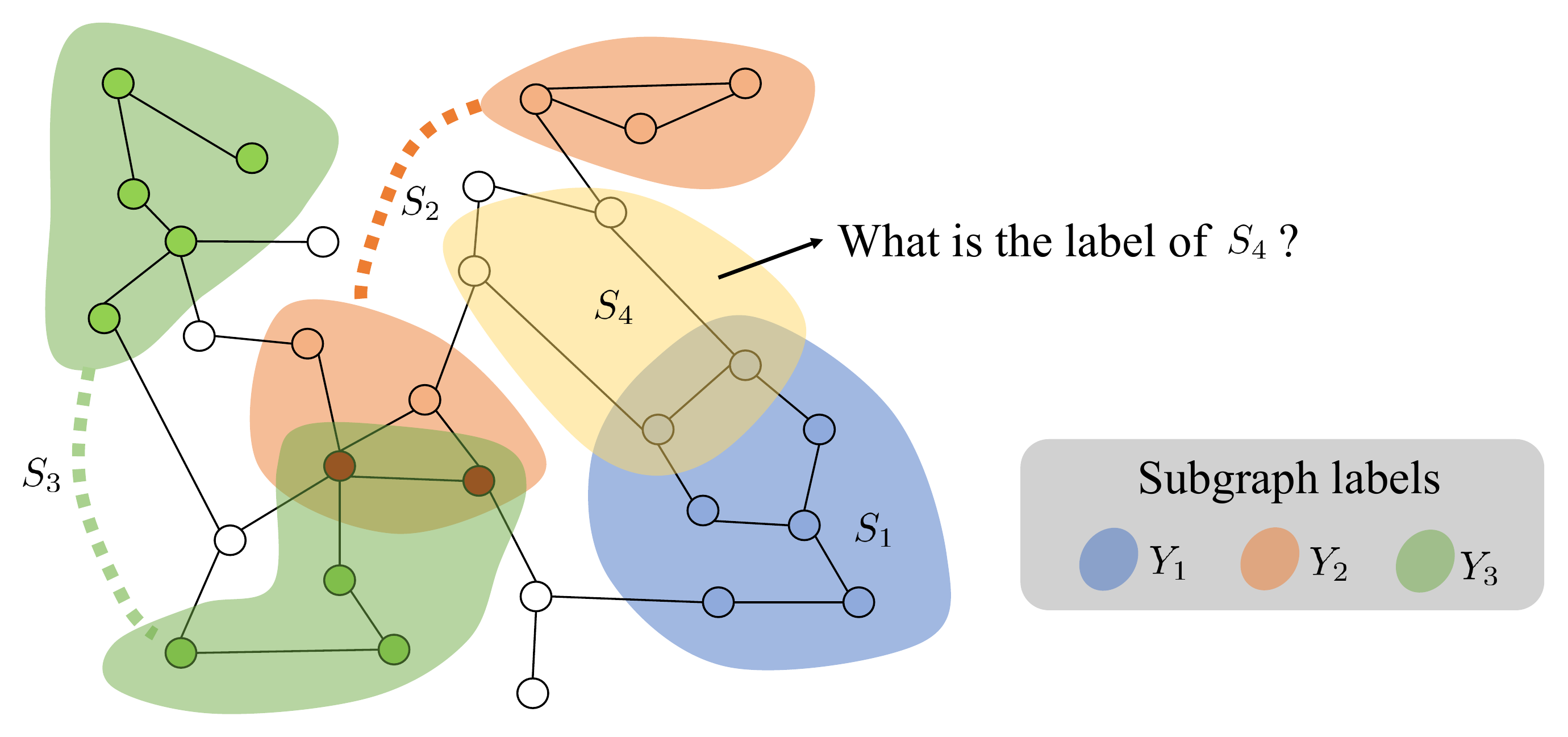} 
\caption{An example of subgraph prediction in base graph $\mathcal{G}$. $\mathcal{S}$ is the set of subgraphs, $\mathcal{S} =\{ S_1, S_2, S_3, S_4\}$, different colors represent different subgraphs. $S_2$ and $S_3$ have multiple isolated components, $S_1$ has only one component. $S_2$ and $S_3$ have shared nodes. $\mathcal{Y}$ is the set of labels, $\mathcal{Y} = \{ Y_1, Y_2, Y_3 \}$. The task is to predict the label of $S_4$.}
\label{figure:subgraph} 
\vspace{0cm}
\end{figure}

The goal of graph representation learning (GRL) \cite{GraphRL} is to learn meaningful low-dimensional vectors for nodes, edges, or entire graphs for downstream tasks \cite{GNN_review_JZ}. Powerful GRLs are usually built on large amounts of training data with supervised signals (labels), but real-world applications face the problem of lacking labeled data. Recent studies notice this challenge in GRL and have proposed some approaches for data-efficient learning on graphs (GEL) \cite{CCSL, HeCo, Megnn, MetaP, Data-efficient-KDD21}. However, existing GEL methods are node-level \cite{DBLP:conf/aaai/0003LNW0S21}, edge-level \cite{SSAL}, or graph-level \cite{GROVER}.
To the best of our knowledge, \textbf{the research on GEL for subgraph prediction is still missing}. Subgraph prediction has important real-world applications such as molecular analysis \cite{zitnik2018biosnap} and clinical diagnostic \cite{bradley2020statistical}. 
Figure \ref{figure:subgraph} illustrates an example of subgraph prediction. There is a base graph and several subgraphs with different labels; subgraph prediction aims to predict whether a particular subgraph has a certain property of interest.  
Subgraph prediction itself is a difficult task because subgraphs are very challenging structures from a topological point of view \cite{SubGNN}. It is necessary to capture the position of all the nodes in the base graph, carry out message passing from nodes inside and outside the subgraph, and consider the dependence of shared edges and nodes in different subgraphs to make the joint prediction.
New challenges arise when subgraph prediction meets ``small'' labeled data: 1) \textbf{How to learn position encoding for nodes in a simple and effective way}? Currently, with limited research on node position encoding, the Position-ware Graph Neural Network (P-GNN) \cite{PGNN} is a feasible solution for node position encoding in subgraph neural networks \cite{SubGNN}. Nevertheless, P-GNN relies on anchors. P-GNN randomly selects nodes on the graph as anchors, then calculates the distance between the target node and anchor nodes, and finally learns the nonlinear distance weighted aggregation on anchors. The overall computational complexity of P-GNN is high.
2) \textbf{How to automatically augment ``good'' subgraphs}? Data augmentation and self-supervised learning are the main ways to solve GEL. The key criterion for data augmentation is to selectively prevent information loss \cite{DBLP:conf/iclr/Xiao0ED21}. Existing graph augmentation methods are mainly based on random perturbation \cite{GraphCL}, sampling \cite{GCC}, or adaptive selection \cite{GCA, JOAO} on nodes and edges. These methods are unsuitable for subgraph augmentation because a subgraph itself contains fewer nodes, richer structural information, and is more susceptible to undesirable perturbations.
3) \textbf{Subgraph GEL faces the ``bias'' problem}. 
We provide statistics of the node coverage in labeled subgraphs of three real-world subgraph datasets\footnote{https://www.dropbox.com/sh/zv7gw2bqzqev9yn/AACR9iR4Ok7f9x1fIAiVCdj3a?dl=0}: HPO-METAB, EM-USER, and HPO-NEURO. 
We find that only a small number of nodes in the base graph are contained in the labeled subgraphs, and the coverage are 17.47\%, 12.41\% and 23.74\%, respectively. 
Details can be found in Table \ref{table:base graph} of Section \ref{sec:datasets}.
Note that shared nodes among subgraphs are not yet counted. 
This leads to the ``bias'' problem which means a small number of nodes dominate the subgraph representation learning. Nodes contained in subgraphs are well trained, while nodes not included in subgraphs fail to be fully learned, resulting in the ``richer get richer'' Matthew effect \cite{bias}. It is a big challenge to the generalization ability of the subgraph prediction problem.

To address the above challenges, in this paper, we propose a \textbf{P}osition-\textbf{A}ware \textbf{D}ata-\textbf{E}fficient \textbf{L}earning (PADEL) framework for subgraph neural networks. 
We design a new position encoding scheme that extends cosine position encoding to Non-Euclidean space to assign a hard phase encoding to each node in the base graph. We apply a random 1-hop breadth-first search to diffuse subgraphs and utilize a variational subgraph autoencoder for data augmentation. We develop exploratory and exploitable views for subgraph contrastive learning. The exploratory view encourages subgraphs to explore unseen nodes in the base graph, and the exploitable view selects internal node connections of the subgraph. After a structure-aware pooling with bidirectional LSTM \cite{bilstm}, the classifier makes the subgraph prediction.
The main contributions of this paper are as follows:
\begin{itemize}[leftmargin=*]
    \item We propose a novel cosine phase position encoding scheme. Our proposed method is anchor-free and is capable of learning an expressive position encoding for each node in the base graph. Compared with existing methods, the computational complexity of our method is reduced.
    \item We propose a new data augmentation and contrastive learning paradigm for data-efficient subgraph representation learning, which alleviates the bias problem in subgraph GEL. 
    \item We conduct extensive experiments on several datasets, and the results show that our proposed method achieves improvements over state-of-the-art methods.
\end{itemize}

\begin{figure*}[t]
	\centering
	\includegraphics[width=0.92\linewidth]{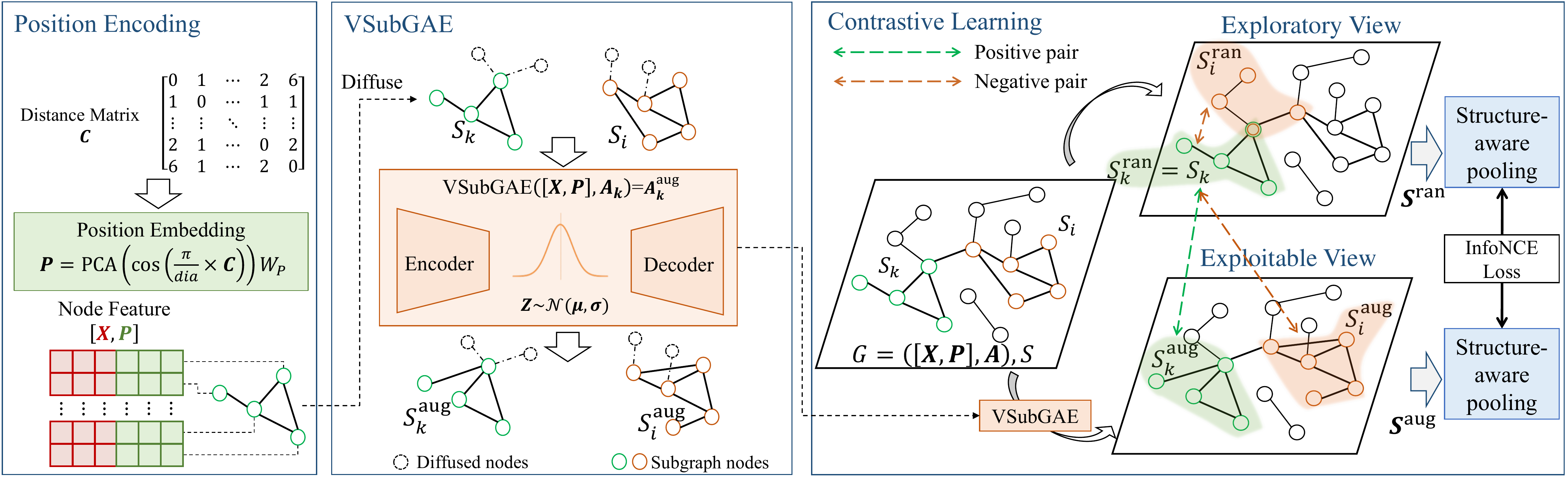}
	\vspace{-0.4cm}
	\caption{The overall architecture and training pipeline of PADEL. $\boldsymbol{C}$ is the symmetric matrix consisting of distance between nodes. $\boldsymbol{P}$ is the position encoding matrix. $\boldsymbol{X}$ is the node embedding matrix. $\boldsymbol{X}$ and $\boldsymbol{P}$ are concatenated to represent nodes' feature. 
    We apply a random 1-hop diffusion for each subgraph and feed it into a variational subgraph autoencoder (VSubGAE) to generate augmented subgraphs. 
	We design exploratory and exploitable views for subgraph contrastive Learning.
	After a structure-aware pooling, we use the InfoNCE loss for training.}
	\label{fig:structure}
	\vspace{-0.4cm}
\end{figure*}

\section{Related Work}
\paragraph{\textbf{Subgraph Neural Networks}} Conventional graph neural networks such as GCN \cite{GCN}, GAT \cite{GAT}, and GIN \cite{GIN} learn node-level or graph-level representations, ignoring the substructures of graphs.
Some recent works note this limitation and propose some improvements. For example, selecting and extracting subgraphs for graph pooling \cite{SUGAR}, identifying important subgraphs to explain graph neural network \cite{SubgraphX}, fusing subgraph hierarchical features for link prediction \cite{SHFF}, classifying subgraphs for link prediction \cite{ARCLink}, and incorporating subgraphs for reasoning \cite{GraIL} and link prediction \cite{Sub-kg-1} on knowledge graphs. 
These efforts focus on exploring and utilizing subgraphs for analysis and inference. However, few studies have paid attention to constructing subgraph neural networks for subgraph prediction (i.e., subgraph classification). 
SPNN \cite{SPNN} learns dependent subgraph patterns for subgraph evolution prediction. SubGNN \cite{SubGNN} is the firstly proposed subgraph neural network that propagates neural messages inside and outside subgraphs in three components: position, neighborhood, and structure. SubGNN follows the position encoding method of P-GNN \cite{PGNN} to select anchors on the base graph randomly. Recently, GLASS \cite{glass} proposes the max-zero-one labeling trick on SubGNN to learn enhanced subgraph representations.
Due to the high computational complexity of SubGNN-related methods, they are not suitable for data-efficient learning.

\paragraph{\textbf{Self-supervised Learning for GEL}} Self-supervised learning for GEL covers node-level GEL, edge-level GEL,  and graph-level GEL. 
For node-level GEL, DGI \cite{DGI} and MVGRL \cite{MVGRL} introduce a discriminator and utilize the global-local contrasting mode. GCA \cite{GCA} proposes an adaptive augmentation framework with intra and inter contrastive views. 
For edge-level GEL, 
SSAL \cite{SSAL} is a self-supervised auxiliary learning method with meta-paths on heterogeneous graphs. SLiCE \cite{SLiCE} incorporates global information and learns contextual node representation. 
For graph-level GEL, GCC \cite{GCC} and GraphCL \cite{GraphCL} are representative graph contrastive learning methods.
CCSL \cite{CCSL} uses unlabeled graphs to pretrain graph encoders and employs a regularizer for data-efficient supervised classification. 
GROVER \cite{GROVER} designs a transformer-style architecture for learning structural information on unlabeled molecular data. 
JOAO \cite{JOAO} proposes an automatic and adaptive data augmentation framework under a bi-level min-max optimization.
InfoGraph \cite{InfoGraph} proposes to encode aspects of data by using different scales of substructures and contrast them with graph-level representations.
GraphLoG \cite{GraphLoG} proposes a base-graph representation learning framework that unifies local instance and global semantic structures. 
To the best of our knowledge, the study of subgraph-level GEL is still blank. 

\section{Methodology}
The overall architecture of our proposed model conforms to the generic self-supervised training paradigm \cite{SSL_review}. PADEL contains components of position encoding, data augmentation, contrastive learning, and pooling. Figure \ref{fig:structure} illustrates the pipeline of PADEL. The downstream task is subgraph prediction which is formulated as follows: $\mathcal{G} = (V, E)$ denotes the base graph, $V$ denotes the node set and $E$ denotes the edge set. $|V|$ and $|E|$ denote the number of nodes and the number of edges, respectively. We focus on undirected graphs. Denote $S_k = (V'_k, E'_k)$ as the $k$-th subgraph of $\mathcal{G}$ if $V_k' \subseteq V$ and $E_k' \subseteq E$. Each subgraph $S_k$ has the corresponding label $Y_k$. A subgraph may compromise multiple isolated components or a single component containing some connected nodes. Given a new subgraph $\hat{S}$ in $\mathcal{G}$, we predict the label of $\hat{S}$.

\subsection{Cosine Phase Position Encoding}

Representing the non-trivial position information for Non-Euclidean structured data (i.e., nodes or subgraphs in the base graph) is still a challenging problem. 
The Position-aware Graph Neural Network (P-GNN) \cite{PGNN} tackles this problem via randomly selecting nodes as anchors and aggregating anchors' information to the target node as position encoding. 
According to the Bourgain theorem \cite{bourgain1985lipschitz}, a graph with $|V|$ nodes requires at least $O(\log^2|V|)$ anchors to guarantee the expressiveness of position encoding. 
SubGNN \cite{SubGNN} inherits the position encoding method of P-GNN. An obvious limitation is the high computational complexity. In each training iteration in SubGNN, it is necessary to re-select anchors randomly, calculate the shortest path from all nodes to anchors and then aggregate anchors' information. Besides, a large space complexity with $O(|V|^2)$ is needed to store the matrix consisting of the shortest paths between nodes in each iteration. The total complexity is very high.  

Another limitation of P-GNN is the expressiveness. The propagation of position information in P-GNN is affected by the distance between target nodes and anchors. The aggregated information will be attenuated if the distance is large; therefore, nodes located in the center of the base graph are easy to obtain strong position signals, while nodes located at the edges of the base graph obtain weak position signals.  
We argue that the above two limitations impede P-GNN for subgraph position encoding. 

To solve the problem, we propose our key insight that \textbf{nodes' position information in the base graph is innate and can be represented by measuring the distance to other nodes}.
Inspired by the sequential position encoding method in Transformer \cite{transformer} and the Distance Encoding Neural Network (DENN) \cite{DEGNN}, we extend the cosine position encoding to Non-Euclidean space to assign a distinctive position encoding for each node in the base graph. 

First, we introduce the concept of ``\emph{diameter}'' to describe the farthest distance of a node to its reachable nodes. 
For the base graph $\mathcal{G} = (V, E)$, let $\boldsymbol{C}$ denote the $|V|\times|V|$ symmetric matrix consisting of distances between nodes in $\mathcal{G}$. For example, $\boldsymbol{C}_{i,j}=\boldsymbol{C}_{j,i}$ denotes the distance (\emph{i.e.}, the length of the shortest path obtained by \emph{Dijkstra's algorithm} \cite{Dijkstra}) between nodes $V_i$ and $V_j$. The diagonal elements of $\boldsymbol{C}$ are all 0, \emph{i.e.}, $\boldsymbol{C}_{i,i}=0$.
If node $V_i$ cannot reach $V_j$, 
$\boldsymbol{C}_{i,j}=\boldsymbol{C}_{j,i}=\infty$. 
We use $dia_i$ to represent the ``diameter'' of the $i$-th node $V_i$,
\begin{equation}
	\label{eq:diameter}
	dia_i = \max_{j} \boldsymbol{C}_{i,j}, \quad \text{where } \boldsymbol{C}_{i,j}\neq \infty, 
\end{equation}
$dia_i$ is the maximum value of the shortest path length from $V_i$ to all reachable nodes. 

Then, we propose to use the cosine function to encode the matrix $\boldsymbol{C}$. Since the value of $dia_i$ could be very large when there are many nodes in graph $\mathcal{G}$, $dia_i$ will fluctuate widely in value. Utilizing the periodic property of the cosine function, we apply the function of Eq.(\ref{eq:phase}) to obtain a scaled matrix $\hat{\boldsymbol{C}}$:   
\begin{equation}
	\label{eq:phase}
	\hat{\boldsymbol{C}}_{i,j}=\left\{
	\begin{aligned}
	    & \cos(\frac{\pi}{dia_i} \times \boldsymbol{C}_{i,j}) & ,&\boldsymbol{C}_{i,j}\neq \infty\\
	    & -1.5 & ,&\boldsymbol{C}_{i,j} = \infty ,
	\end{aligned}
	\right.
\end{equation}
where $\cos(\frac{\pi}{dia_i} \times \boldsymbol{C}_{i,j})$ means for each node $v_i$, its all reachable nodes $v_j$ (including $v_i$ itself) are mapped to phase $[0, \pi]$.  In other words, among all reachable nodes, the distance of the node closest to $v_i$ (itself) is mapped to 1. The distance of other reachable nodes is mapped to smaller values, and if the distance is farther, the mapped value is closer to -1. Unreachable nodes are mapped to a smaller number than -1 (\emph{e.g.}, -1.5).
If we think of each element $\boldsymbol{C}_{i,j}$ in $\boldsymbol{C}_i$ as a discrete phase value of the cosine function, $\boldsymbol{C}_i$ can be viewed as the superposition of different cosine phase values. 
The processed matrix $\hat{\boldsymbol{C}}$ has the following properties:
\begin{itemize}[leftmargin=*]
	\item \textbf{Normalization}: The computed phases $\frac{\pi}{dia_i} \times \boldsymbol{C}_{i,j}$ in Eq. (\ref{eq:phase}) range in $[0, \pi]$, so $\hat{\boldsymbol{C}}_{i,j} \in [-1,1]$ for reachable nodes, and -1.5 for the others.
	\item \textbf{Distinctiveness}: Each row in $\hat{\boldsymbol{C}}$ is distinguished from others, because $\boldsymbol{C}$ is symmetric and $\hat{\boldsymbol{C}}_{ii} = 1$.
	\item \textbf{Interpretability}: Each element in $\hat{\boldsymbol{C}}_i$ is closely related to the minimal number of hops between node $i$ and all nodes.
\end{itemize}

Next, we employ the Principal Component Analysis (PCA) \cite{PCA} to reduce the dimension of $\hat{\boldsymbol{C}}$ to the size of $|V|\times d'$, where $d'$ is a manually set hyperparameter, 
\begin{equation}
\label{eq:pca}
    \overline{\boldsymbol{C}}= \text{PCA} (\hat{\boldsymbol{C}}).
\end{equation}

Last, we perform a linear projection on $\overline{\boldsymbol{C}}$ to generate the position embedding matrix $\boldsymbol{P}$ with the size of $|V|\times d$,
\begin{equation}
\label{eq:pe}
    \boldsymbol{P} =  \overline{\boldsymbol{C}}\boldsymbol{W}_P,
\end{equation} 
where $\boldsymbol{W}_P \in \mathbb{R}^{d' \times d}$ is a learnable parameter matrix.
The reason we introduce the projection matrix $P$ is to make position encoding scalable. Since we need to manually specify the dimension for PCA, it’s not flexible if we want to adjust the dimension of position encoding. If we apply the projection matrix $\boldsymbol{W}_P$ directly to $\hat{\boldsymbol{C}}$, $\boldsymbol{W}_P$ will be large, and more importantly, it brings instability to the subsequent training of the model (i.e., the KL term in Eq. (\ref{eq:VGAE}) will be extremely large).

Compared with P-GNN, our proposed cosine phase position encoding has the following advantages: 1) Our method is anchor-free. The position information is measured only by distance between nodes. 2) The Cosine phase position encoding matrix $\hat{\boldsymbol{C}}$ has the nature of normalization, distinctiveness, and interpretability as analyzed above. 
3) Our position encoding method is expressive. We provide a case study in section~\ref{case_study} (See Fig. \ref{fig:PE}).
The above three advantages make our method not only a general-purpose position encoding method but also suitable for data-efficient subgraph neural networks. 



\subsection{Variational SubGraph AutoEncoder for Data Augmentation}
It is essential to prevent information loss during data  augmentation \cite{DBLP:conf/iclr/Xiao0ED21}. The quality of an augmented graph is crucial in graph contrastive learning \cite{JOAO}. Existing methods \cite{GraphCL, GCC, GCA, JOAO} are not suitable for subgraph augmentation because a subgraph contains only a small part of the nodes in the base graph, but contains rich position and structural information, which needs to be augmented with appropriate strategies.
Unlike existing graph augmentation methods, we utilize generative models to augment subgraphs in PADEL. Specifically, we design a Variational SubGraph AutoEncoder (VSubGAE) with random 1-hop node diffusion. 

Given the base graph $\mathcal{G}$ with $|V|$ nodes, suppose the node embedding matrix is $\boldsymbol{X} \in \mathbb{R}^{|V| \times d}$.
We assign a unique feature vector to each node since some nodes may be shared in different subgraphs, and some nodes may not appear in any subgraph.   
For the $k$-th subgraph $\boldsymbol{S}_k$, its corresponding adjacency matrix is $\boldsymbol{A}_k$.

\paragraph{\textbf{The Encoder.}} 
For the $i$-th node $v_i\in S_k$, we have the corresponding node feature vector $\boldsymbol{x}_i$ and the position encoding $\boldsymbol{p}_i$. We concatenate $\boldsymbol{x}_i$ and $\boldsymbol{p}_i$, and feed them into the encoder of VSubGAE which is a two-layer GCN:
\begin{equation}
    \label{eq:hidden}
    q(\boldsymbol{z}_{k,i}|[\boldsymbol{X}, \boldsymbol{P}],\boldsymbol{A}_k)\sim \mathcal{N}(\boldsymbol{\mu}_{i}([\boldsymbol{X}, \boldsymbol{P}],\boldsymbol{A}_k),\boldsymbol{\sigma}_i([\boldsymbol{X}, \boldsymbol{P}],\boldsymbol{A}_k)),
\end{equation}
where $[\cdot, \cdot]$ denotes concatenation, $\mathcal{N}(\cdot,\cdot)$ denotes the Gaussian distribution. $\boldsymbol{\mu}=\mathrm{GCN}_{\mu}([\boldsymbol{X}, \boldsymbol{P}],\boldsymbol{A}_k)$ and $\boldsymbol{\sigma}=\mathrm{GCN}_{\sigma}([\boldsymbol{X}, \boldsymbol{P}],\boldsymbol{A}_k)$. $\mathrm{GCN}_{\mu}$ and $\mathrm{GCN}_{\sigma}$ are defined as:
\begin{equation}
    \begin{aligned}
    &\mathrm{GCN_{\mu,\sigma}}([\boldsymbol{X}, \boldsymbol{P}],\boldsymbol{A}_k)=\boldsymbol{\tilde{A}}_k\mathrm{ReLU}(\boldsymbol{\tilde{A}}_k[\boldsymbol{X}_{v\in V_k},\boldsymbol{P}_{v\in V_k}]\boldsymbol{W}_1)\boldsymbol{W}_{\mu,\sigma}, 
    \end{aligned}
\end{equation}
where $\boldsymbol{\tilde{A}}_k=\boldsymbol{D}_k^{-\frac{1}{2}}\boldsymbol{A}_k\boldsymbol{D}_k^{-\frac{1}{2}}$ and $\boldsymbol{D}_k$ is the degree matrix of $S_k$. $\boldsymbol{W}_1 \in \mathbb{R}^{2d \times 2d}$ is the shared learnable matrix, $\boldsymbol{W}_{\mu}, \boldsymbol{W}_{\sigma} \in \mathbb{R}^{2d \times 2d}$ are learnable matrices.
The parameters of GCN layers are shared among all of the subgraphs. The inference process is defined as:
\begin{equation}
    \label{eq:infer}
    q(\boldsymbol{Z}|[\boldsymbol{X}, \boldsymbol{P}],\mathcal{A})=\prod_{k=1}^K\prod_{i=1}^{|V'_k|}q(\boldsymbol{z}_{k,i}|[\boldsymbol{X}, \boldsymbol{P}],\boldsymbol{A}_k),
\end{equation}
where $\mathcal{A}=\{\boldsymbol{A}_1,\boldsymbol{A}_2,\cdots,\boldsymbol{A}_K\}$ is the collection of adjacency matrices with $K$ elements. The reparameterization trick is used to sample a node feature: $\boldsymbol{z}_{k,i} = \boldsymbol{\mu}_{i}([\boldsymbol{X}, \boldsymbol{P}],\boldsymbol{A}_k) + \boldsymbol{\sigma}_i([\boldsymbol{X}, \boldsymbol{P}],\boldsymbol{A}_k) \odot \boldsymbol{\epsilon}$, where $\odot$ represents point-wise product between two vectors, $\boldsymbol{\epsilon} \sim \mathcal{N}(0,\boldsymbol{I}_k)$. 

\paragraph{\textbf{The decoder.}} We use a simple inner product between latent variables as the decoder, which obtains the possibility of two edges connecting to each other. Thus we have:
\begin{equation}
    \begin{aligned}
    \label{eq:gen}
    p(\mathcal{A}|\boldsymbol{Z}) &=\prod_{k=1}^K\prod_{i=1}^{|V_k'|}\prod_{j=1}^{|V_k'|}p(A_{k,ij}|\boldsymbol{z}_{k,i},\boldsymbol{z}_{k,j})\\
    &=\prod_{k=1}^K\prod_{i=1}^{|V_k'|}\prod_{j=1}^{|V_k'|}\mathrm{Sig}(\boldsymbol{z}_{k,i}\boldsymbol{z}_{k,j}),
\end{aligned}
\end{equation}
where $\mathrm{Sig}(\cdot)$ denotes the Sigmoid function.

\paragraph{\textbf{VSubGAE Optimization.}} Following VGAE \cite{VGAE} and $\beta$-VAE \cite{betaVAE}, the optimization goal of VSubGAE is to maximize the variational Evidence Lower Bound (ELBO) with coefficient $\beta$ , which balances the reconstruction accuracy of subgraphs with the independence constraints of the latent variables (\textit{i.e.} the KL divergence term in $\mathcal{L}_{VSubGAE}$):
\begin{equation}
    \label{eq:VGAE}
    \begin{aligned}
        \mathcal{L}_{VSubGAE } = & \mathbb{E}_{\boldsymbol{Z}\sim q(\boldsymbol{Z}|[\boldsymbol{X}, \boldsymbol{P}],\mathcal{A})}\log{p(\mathcal{A}|\boldsymbol{Z})}\\
        & -\beta\cdot\mathrm{KL}(q(\boldsymbol{Z}|[\boldsymbol{X}, \boldsymbol{P}],\mathcal{A})||p(\boldsymbol{Z})),
    \end{aligned}
\end{equation}
where $p(\boldsymbol{Z})\sim\mathcal{N}(0,\boldsymbol{I})$ is the prior distribution of latent variables $\boldsymbol{Z}$, following the common practice in \cite{betaVAE,VGAE}.



\subsection{Contrastive Learning in PADEL}
We train PADEL in a self-supervised manner by contrasting positive and negative sample pairs. As we analyzed in section 1, nodes in subgraphs are only a small part of the base graph. This results in a bias problem which means ``hot'' nodes (contained in subgraphs) will dominate model training, and ``long-tail'' nodes (not included in subgraphs) fail to be fully learned. 
Although we alleviate the problem to some extent by using 1-hop diffusion in the VSubGAE component, nodes far from subgraphs are still unexplored. To further alleviate the bias problem, we propose Exploratory and Exploitable views for subgraph contrastive learning, which is the right part in Figure \ref{fig:structure}.

\paragraph{\textbf{The Exploratory View}}
\label{para:explore}
We design an exploratory view (abbreviated as Explore-View) for data augmentation to enable our model to explore unseen nodes in the base graph. We randomly sample subgraphs in $\mathcal{G}$ as augmented data, which allows our model to \emph{explore} more distant undetected nodes.
Let the $k$-th subgraph be the positive sample, we generate exploratory subgraphs $\mathcal{S}^{\text{ran}}$ by:
\begin{equation}
    \label{eq:random}
    S^{\text{ran}}_i=\left\{
    \begin{aligned}
    	&S_k, & \text{if } i=k, \\
    	&\text{RandomWalk}(hop), &  \text{otherwise},
    \end{aligned}
    \right.
\end{equation}
where $\text{RandomWalk}(hop)$ means performing $hop$-step random walks on the base graph to generate a random subgraph. $hop$ is equal to the average number of nodes for a given dataset.


\paragraph{\textbf{The Exploitable View}}
In addition to exploring distant nodes in the base graph, we also need to find the appropriate perturbation from the subgraph for augmentation adaptively.
We consider this an exploitable view (abbreviated as Exploit-View) that changes a subgraph's internal connections.
To augment data adaptively, we use VSubGAE to generate the augmented adjacency matrix $A_k^{aug}$ for subgraph $S_k^{aug}$ by Eq. (\ref{eq:infer}) and (\ref{eq:gen}), which can be denoted as:
\begin{equation}
	\boldsymbol{S}_k^{aug} \leftarrow  \boldsymbol{A}^{\text{aug}}_k = \text{VSubGAE}([\boldsymbol{X}, \boldsymbol{P}], \boldsymbol{A}_k).
\end{equation}

\paragraph{\textbf{Contrastive Learning}}
Given the subgraph set $\mathcal{S}$ containing $K$ subgraphs, we train all $K$ subgraphs in one training step. 
Each subgraph is treated as a positive sample, while others are treated as negative ones. Exploratory subgraphs are sampled for each positive sample separately in each training step. 

The Exploratory view augmentation and Exploitable view augmentation generate augmented subgraphs  $\mathcal{S}^{\text{ran}}$ and $\mathcal{S}^{\text{aug}}$, respectively. 
For the $k$-th subgraph, we choose the exploratory view $\boldsymbol{S}^{\text{ran}}_k$ and the exploitable view $\boldsymbol{S}^{\text{aug}}_k$ as the positive pair.
We define positive and negative pairs as follows:
\begin{itemize}
    \item Positive pair: $(S^{\text{ran}}_k,S^{\text{aug}}_k)$.
    \item Explore-View negative pairs: $\{(S^{\text{ran}}_k, S^{\text{ran}}_i)$, where $i\neq k\}$.
    \item Exploit-View negative pairs: $\{(S^{\text{ran}}_k, S^{\text{aug}}_i)$, where $i\neq k\}$.
\end{itemize}

The InfoNCE loss is computed as follows:
\begin{equation}
\label{eq:infonce}
	\begin{aligned}
		&\mathcal{L}_{\text{InfoNCE}}(k,\boldsymbol{E}^{\text{ran}} ,\boldsymbol{E}^{\text{aug}} )= \\
		& - \log \frac{e^{\theta\left(S^{\text{ran}}_k ,S^{\text{aug}}_k\right)}}{\underbrace{e^{\theta\left(S^{\text{ran}}_k ,S^{\text{aug}}_k\right) }}_{\text {Positive Pair }}+\underbrace{\sum_{i \neq k} e^{\theta\left(S^{\text{ran}}_k, S^{\text{ran}}_i\right)}}_{\text {Explore-View negative pairs }}+\underbrace{\sum_{i \neq k} e^{\theta\left(S^{\text{ran}}_i, S^{\text{aug}}_k\right)}}_{\text {Exploit-View negative pairs }}},
	\end{aligned}
\end{equation}
where $\theta$ is the contrastive score as the sum of cosine similarity for both neighbor-position encoding $\boldsymbol{E}^\text{NP}$ and structure-position encoding $\boldsymbol{E}^\text{S}$ (See section \ref{Sec:pooling}):
\begin{equation}
\label{eq:sim}
    \theta(S_i, S_j) = \frac{\left<\boldsymbol{E}^\text{NP}_i, \boldsymbol{E}^\text{NP}_j\right>}{||\boldsymbol{E}^\text{NP}_i||\times||\boldsymbol{E}^\text{NP}_j||}+\frac{\left<\boldsymbol{E}^\text{S}_i, \boldsymbol{E}^\text{S}_j\right>}{||\boldsymbol{E}^\text{S}_i||\times||\boldsymbol{E}^\text{S}_j||},
\end{equation}
where $\left< \cdot , \cdot \right>$ denotes inner product and $||\cdot||$ denotes the norm.

Next section will discuss how to compute subgraph encoding $\boldsymbol{E}^\text{NP}, \boldsymbol{E}^\text{S}$ with our structure-aware subgraph pooling model.

\subsection{Structure-Aware Subgraph Pooling} 
\label{Sec:pooling}
\begin{figure}
	\centering
	\includegraphics[width=\linewidth]{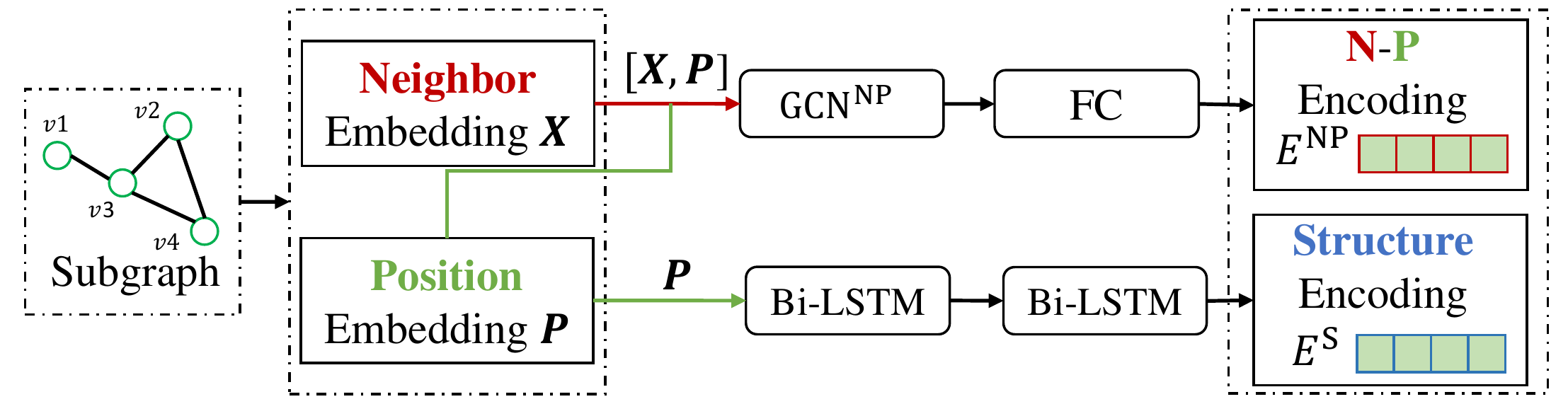}
	\caption{The structure-aware subgraph pooling component. The lower part illustrates how Bi-LSTM layers extract structure encoding $E^{NP}$ of a given subgraph.}
	\label{fig:subpooling}
\end{figure}

Graph readout plays an important role in GEL \cite{RethinkPooling}. We develop a structure-aware subgraph pooling method. The pooling architecture is shown in Figure \ref{fig:subpooling}. Our pooling component is designed to learn structure-aware subgraph representations that can capture three aspects of information: neighborhood, position, and structure. The inputs are node embedding matrix $\boldsymbol{X}$ and position encoding matrix $\boldsymbol{P}$. We implement structure-aware subgraph pooling through neighbor nodes' message passing with position information and structure extraction based on the position information.

\paragraph{\textbf{Neighbor Aggregation with Position Information}}

Node embedding and position encoding of subgraph $S_k$ are concatenated as the input of a GCN layer, and then through a fully-connected layer followed by an average pooling layer: 
\begin{equation}
	\boldsymbol{E}_k^\text{NP} = avg\_pool( \text{ReLU}(\text{GCN}^\text{NP}([\boldsymbol{X},\boldsymbol{P}], \boldsymbol{A}_k)\boldsymbol{W}^\text{FC})),
\end{equation} 
where $[\cdot, \cdot]$ denotes concatenation, $\boldsymbol{A}_k$ is the adjacency matrix of the input subgraph $\boldsymbol{S}_k$, $\boldsymbol{W}^\text{FC} \in \mathbb{R}^{2d \times \hat{d}}$ is the learnable parameter matrix.


\paragraph{\textbf{Structure Extraction with Position Information}}
We believe that the structural information in the graph should be innate and task-agnostic.
In subgraphs, the encoded position encoding matrix $\boldsymbol{P}$ is considered as the prior information describing each node's position in the base graph. Naturally, we
extract the subgraph's structural feature denoted as $\boldsymbol{E}^\text{S}$ from its nodes' position information.
We adopt a two-layer Bi-LSTM \cite{bilstm} as the extractor. Since Bi-LSTM is capable of learning order-invariant information, it is useful to extract the subgraph's structural information \cite{SubGNN}. Nodes in the subgraph are reorganized into a sequence and fed into the Bi-LSTM unit.
We use the hidden state of the last Bi-LSTM layer as the structure encoding:
\begin{equation}
 \boldsymbol{E}_k^\text{S} = \text{ReLU}(\text{Bi-LSTM}(\boldsymbol{P}_{v\in V_k})).
\end{equation}
We also tried different kinds of layers, such as attention layers and one Bi-LSTM layer; we found that using two Bi-LSTM layers is the best choice.

\subsection{Objective Functions and Optimization}
For subgraph encoding $\boldsymbol{E}=\{\boldsymbol{E}^\text{NP}, \boldsymbol{E}^\text{S}\}$, we use a 1-layer fully-connected networks to compute classification logits $\hat{\boldsymbol{Y}}$ as follows:
\begin{equation}
\label{eq:class}
    \hat{\boldsymbol{Y}} = (\boldsymbol{E}^\text{NP} + 
    \boldsymbol{E}^\text{S}) \boldsymbol{W}^{\text{S}},
\end{equation}
where $\boldsymbol{W}^{\text{S}} \in \mathbb{R}^{\hat{d} \times l}$ is the parameter matrix, $l$ depends on the downstream task. $\hat{\boldsymbol{Y}}$ is a vector with the same number of elements as the total class number. Let $Y$ denote the one-hot vector of the ground-truth label, 
we use the Cross-Entropy Loss $\mathcal{L}_{\text{CE}}$ as follows:
\begin{equation}
    \label{eq:ce}
    \mathcal{L}_{\text{CE}}(Y, \hat{\boldsymbol{Y}}) = -\sum_{i} Y_i\log \frac{exp^{\hat{Y}_i}}{\sum_j exp^{\hat{Y}_j}}.
\end{equation}

\section{Experiments}
We provide the pseudocode of the algorithm of VSubGAE’s random 1-hop subgraph diffusion, the training pipeline of PADEL, and the training data in our anonymous external repository\footnote{https://github.com/AlvinIsonomia/PADEL/}.
\subsection{Datasets}
\label{sec:datasets}

\begin{table}[]\small
\vspace{-0.4cm}
\caption{Statistics of Datasets.}\label{table:base graph}
\vspace{-0.4cm}
\centering
\resizebox{\linewidth}{!}{
{
\begin{tabular}{@{}l|c|c|c|c|c|c@{}}
\toprule
Dataset   & \#Nodes & \#Edges & \#Subgraphs & \makecell{\#Nodes in \\Subgraphs} & Coverage  & Multi-Label \\ \midrule
HPO-METAB & 14587   & 3238174 & 2400   &2548 & 17.47\%   &           \\
EM-USER   & 57333   & 4573417 & 324    &7115 & 12.41\%    &           \\
HPO-NEURO & 14587   & 3238174 & 4000   &3463 & 23.74\%       & Y         \\ \bottomrule
\end{tabular}
}
}
\vspace{-0.5cm}
\end{table}
To evaluate the effectiveness of our proposed method PADEL, we conduct extensive experiments on three real-world datasets: HPO-METAB, EM-USER, and HPO-NEURO. All three datasets are released in \cite{SubGNN} by the Harvard team and are used in the source code of SubGNN \footnote{https://github.com/mims-harvard/SubGNN}. 
Table \ref{table:base graph} describes their statistics. 
\begin{itemize}[leftmargin=*]
\item \textbf{HPO-METAB} is a dataset of clinical diagnostic task for rare metabolic disorders. The base graph contains information on phenotypes and genotypes of rare diseases, in which the nodes denote the genetic phenotypes of the diseases and the edges represent the relationships between phenotypes. Information on relationships is obtained from the Human Phenotype Ontology (HPO) \cite{Khler2019ExpansionOT} and rare disease diagnostic data \cite{Austin2017MedicalRN,Kim2011GeneticCN,Hartley2020NewDA}. Each subgraph consists of a set of phenotypes associated with rare monogenic metabolic diseases, with a total of 6 metabolic diseases.
Subgraph labels represent the diagnosis categories.

\item \textbf{HPO-NEURO} is used to diagnose rare neurological diseases. It shares the same base graph with HPO-METAB but has different subgraphs. Each subgraph has multiple disease category labels and contains 10 neurological disease categories.

\item \textbf{EM-USER}'s base graph is from the Endomondo \cite{Endomondo} social fitness network used to analyze user properties, where nodes represent exercises, and edges will exist if users finish exercises. Subgraphs show the user's exercise records, while labels represent the 2 genders of the user.
\end{itemize}

\subsection{Experimental settings}
\paragraph{\textbf{Data Processing.}}Following SubGNN \cite{SubGNN}, for HPO-METAB and HPO-NEURO datasets, we randomly divide 80\% of the data into the training set, 10\% of the data into the validation set, and 10\% of the data into the test set. For the EM-USER dataset, the proportions are 70\%, 15\%, and 15\%, respectively. Since our model is for \emph{data-efficient} learning, we \textbf{randomly select 10\% of the original training set as our data-efficient training set. All methods are trained with the same data-efficient training set}. The validation and test sets remain the same with the original dataset.

\paragraph{\textbf{Baselines}}
We compare the performance of PADEL with the following baseline methods.
\textbf{GCN} \cite{GCN} is the conventional graph convolutional network \cite{GCN}. 
\textbf{GAT} \cite{GAT} is the graph neural network with attention mechanism. 
\textbf{GIN} \cite{GIN} is the graph isomorphism network with powerful representation learning capabilities. 
\textbf{GraphCL} \cite{GraphCL} is the representative graph contrastive learning method.
\textbf{GCA} \cite{GCA} is a node-level graph contrastive learning method with adaptive augmentation. 
\textbf{JOAO} \cite{JOAO} is the automated graph augmentation method which augments graphs adaptively, automatically and dynamically.
\textbf{SubGNN} \cite{SubGNN} and \textbf{GLASS} \cite{glass} are the state-of-the-art (SOTA) subgraph neural networks. 

\paragraph{\textbf{Evaluation Protocals}}
We adopt the \emph{Micro F1 Score}\cite{SubGNN} as the evaluation metric. Higher scores indicate better performance.

\paragraph{\textbf{Reproducibility Settings}}
We develop our model with PyTorch \cite{Pytorch} and NetworkX \cite{NetworkX}. 
All methods are trained on a single GeForce RTX 2080Ti GPU. \emph{We repeat the experiment 10 times for all methods, each time with a different random seed}. After ten experiments, we record the mean and standard deviation of the results.
For a fair comparison, the input dimension for all methods is set to 64. In our model, the dimension of the node feature is set to 32, and the dimension of position embedding is also set to 32, so the total dimension is 64. We set the batch size to 32 and set the maximal training epoch to 3000 for all methods to ensure training convergence. 
We use the AdamW \cite{AdamW} optimizer for optimization and set the weight decay to 1e-2 in AdamW. We search the learning rate in the range of \{1e-3, 5e-3, 1e-2\} for all methods. The coefficient $\beta$ in VSubGAE is set to 0.2 according to beta-VAE \cite{betaVAE}.

For GCN and GAT, we use a default 3-layer graph neural network. For GIN, we use a default 3-layer perceptron. For GraphCL \cite{GraphCL} and JOAO \cite{JOAO}, we convert test datasets to the TUDataset format \footnote{https://chrsmrrs.github.io/datasets/docs/format/} while retaining all subgraph internal edges and subgraph labels, and we use the default semi-supervised learning setup (the scaling parameter is set to 4 for HPO-METAB and HPO-NEURO datasets, and 5 for the EM-USER dataset). Since GCA \cite{GCA} is for node classification, we extend it to the subgraph classification task by adding an average pooling layer. For SubGNN \cite{SubGNN}, we use the optimal model hyperparameters suggested in the official source code \footnote{ https://github.com/mims-harvard/SubGNN/tree/main/best\_model\_hyperparameters}. 

For multi-label classification on the HPO-NEURO dataset, we treat it as a multiple binary classification task. We calculate the accumulated loss on each label position by using the Binary Cross Entropy loss:
\begin{equation}
    \mathcal{L}_{\text{BCE}} = -\sum_{i} [Y_i\log({\text{Sig}(\hat{Y}_i)}) + (1-Y_i)\log(1-\text{Sig}(\hat{Y}_i))],
\end{equation}
where Sig denotes the Sigmoid function. 

\begin{table}[]\small
\vspace{-0.4cm}
\caption{Performance on GEL on 10\% original training set (the best in bold). Underline denotes the best baseline.}\label{tab:all performance}
\vspace{-0.45cm}
\centering
\begin{tabular}{@{}l|c|c|c@{}}
\toprule
Method     & \makecell{HPO-METAB \\ (10\%)}  & \makecell{EM-USER \\ (10\%)}  & \makecell{HPO-NEURO \\ (10\%)}  \\ \midrule
GCN        & 24.43±1.61 & 48.37±0.93 & 44.84±1.04 \\
GAT        & 27.06±3.50 & 51.63±5.58 & 36.82±2.71 \\
GIN        & 11.91±9.89 & 45.58±6.34 & 33.24±6.60 \\
GraphCL    & 24.00±0.38 & 55.17±2.32 & 32.01±0.36 \\
GCA        & 35.32±3.47 & 58.18±6.46 & 48.28±1.72 \\
JOAO       & 24.16±0.40 & 54.83±1.22 & 32.45±0.35 \\
SubGNN     & {\ul 37.87±4.02} & {\ul 58.78±6.63} & {\ul 51.36±2.09} \\
GLASS     & 31.58±6.67 & 53.46±5.06 & 47.68±4.19 \\
Our Method & \textbf{44.72±1.34} & \textbf{68.45±5.51} & \textbf{53.33±0.75} \\ \midrule
Improvement (\%)& 18.09 & 16.45 & 3.84\\ \bottomrule
\end{tabular}
\vspace{-0.4cm}
\end{table}

\begin{table*}\small
\vspace{-0.4cm}
\centering
\caption{The performances on three datasets with different scales of training data (the best in bold).}
\vspace{-0.4cm}
\label{tab:data-efficient performance}
\begin{tabular}{lccccccl}

\bottomrule
                           & 10\%       & 20\%       & 30\%       & 40\%       & 50\%       & 100\%      &            \\ \midrule 
\multirow{3}{*}{HPO-METAB} & 37.87±4.02 & 42.17±3.64 & 46.51±3.00 & 48.13±3.71 & 50.04±2.76 & 54.72±2.48 & SubGNN     \\
                           & 31.58±6.67 & 44.88±5.14 & 46.80±4.72 & 51.64±2.55 & \textbf{57.02±5.26} & \textbf{62.98±3.84} & GLASS \\
                           & \textbf{44.72±1.34} & \textbf{49.33±1.47} & \textbf{49.69±1.40} & \textbf{51.64±1.62} & 55.85±1.66 & 56.26±0.91 & Our Method \\ \hline
\multirow{3}{*}{EM-USER}   & 58.78±6.63 & 68.98±4.54 & 73.27±7.27 & 80.41±1.00 & 82.45±1.63 & 81.43±4.87 & SubGNN     \\
                           & 53.46±5.06 & 57.12±4.65 & 61.66±8.40 & 76.74±6.88 & 82.44±2.08 & 84.92±3.56 & GLASS \\
                           & \textbf{68.45±5.51} & \textbf{73.56±5.30} & \textbf{77.33±3.56} & \textbf{82.00±2.89} & \textbf{84.00±2.95} & \textbf{84.96±3.46} & Our Method \\ \hline
\multirow{3}{*}{HPO-NUERO} & 51.36±2.09 & 56.63±1.65 & 59.07±1.15 & 60.04±0.93 & 60.11±1.82 & 64.20±1.63 & SubGNN     \\
                           & 47.68±4.19 & 53.64±1.41 & \textbf{61.18±2.90} & \textbf{63.20±0.72} & \textbf{64.44±1.48} & \textbf{68.42±0.66} & GLASS \\
                           & \textbf{53.33±0.75} & \textbf{58.41±1.35} & 60.35±1.20 & 61.09±1.02 & 62.70±0.87 & 65.18±1.41 & Our Method \\ \bottomrule
\end{tabular}
\vspace{-0.4cm}
\end{table*}

\subsection{Performance Comparison}
\paragraph{\textbf{Overall Performance}.} Table \ref{tab:all performance} describes the overall performance, we can find the following observations:
\begin{itemize}[leftmargin=*]
    \item PADEL outperforms all baseline methods on three datasets, and the improvement is achieved. PADEL achieves an average improvement of 12.79\% compared with the SOTA method SubGNN.
    \item Conventional methods GCN, GAT, and GIN don't perform well. The possible reason is that their representation learning ability decreases sharply in subgraph neural networks.
    \item Graph augmentation-based methods GraphCL, GCA, and JOAO perform better than conventional methods in general, but the results are subtle. We find that GraphCL and JOAO perform worse on HPO-METAB and HPO-NEURO datasets than GCN and GAT. 
    It reflects the fact that existing graph-level augmentation methods are very limited in subgraph neural networks. 
    GCA performs much better than GraphCL and JOAO on three datasets. The possible reason is that GCA is a node-level self-supervised learning method, it learns node representation effectively by the intra- and inter-view contrastive learning using adaptive augmentation.
    \item SubGNN achieves the previous SOTA performance. SubGNN is a sophisticated method for extracting subgraph representations and passing messages between subgraphs, and this is the reason why SubGNN performs best of all baseline methods. Our proposed model PADEL outperforms SubGNN by large margins. There are three possible reasons: 1) Nodes' position information is well captured and learned. 2) The augmentation-contrastive learning paradigm in PADEL is effective. 3) The pooling method in PADEL captures subgraphs' structures.   
\end{itemize}
\paragraph{\textbf{Performance on different scales of data}.}We compare PADEL with SOTA baselines SubGNN and GLASS by taking 10\%, 20\%, 30\%, 40\%, 50\%, and 100\% of the training set on three datasets to verify their performance. Table \ref{tab:data-efficient performance} describes the results. We have the observation that PADEL outperforms SubGNN across the board. PADEL shows comparative performance with GLASS, but outperforms GLASS in most data-efficient situations.

\paragraph{\textbf{Time Cost}.}To compare the time cost, we test on the HPO-METAB (10\%) dataset using a single GPU. SubGNN requires preprocessing of the data file, spending 12 minutes calculating the metric of shortest path length, 4 hours calculating the metric of subgraph similarity, and 12 hours calculating node embedding. The training time of SubGNN is 227 seconds for 100 epochs.
\begin{table}[H]\small
\vspace{-0.37cm}
\caption{The time cost on HPO-METAB (10\%) dataset.}
\vspace{-0.4cm}
\setlength{\tabcolsep}{1mm}
\label{tab:time}
\begin{tabular}{lccccc}
\bottomrule
   & \textbf{Step}                   & \textbf{SubGNN}              & \textbf{GLASS}     & \textbf{PADEL}         & \textbf{Speed Up}   \\ \midrule 
\multirow{2}{*}{Pretrain} &  Metrics                   & 4 h 12 min           & 0              & 12 min                       &\multirow{2}{*}{\textbf{46 / 65} $\times$} \\ \cline{2-5}
                                 &  Embedding                & 12 h    & 23 h  & 557 sec &                     \\ \midrule 
Train        & Subgraph                     & 227 sec                         & 79 sec  & 39 sec                 & \textbf{6 / 2} $\times$ \\ 
\bottomrule
\end{tabular}
\vspace{-0.45cm}
\end{table}
In the pre-training phase, GLASS doesn't need to compute metrics, but it need 23 hours to pretrian its node embeddings. The average training time of GLASS is 79 seconds for 100 epochs.
In comparison, PADEL spends 12 minutes calculating the shortest path lengths, 17 seconds calculating VSubGAE, and 540 seconds on contrast learning. The training time of PADEL is 39 seconds for 100 epochs. Compared with SuhbGNN, the pretraining time of PADEL is 46 times shorter, and the training time is 6 times shorter. 
PADEL also takes 65 $\times$ time shorter in pretraining and 2 $\times$ time shorter in training than GLASS.
Results are shown in Table \ref{tab:time}.
We find that the most time-consuming step in PADEL is calculating the distance matrix $\boldsymbol{C}$ for 12 minutes because of the $O(|V|^3)$ time complexity of the \emph{Dijkstra's algorithm}. Although existing SubGNN and its variants \cite{SubGNN,glass} also need to pre-compute $\boldsymbol{C}$, we leave the problem of learning position embedding more efficiently and making it suitable for larger-scale graphs in future works.


\subsection{Ablation Study}
To investigate different components' effectiveness in PADEL, we design eight cases for ablation study ($C_0$-$C_7$).
$C_0$: Only apply the \textbf{S}ubgraph \textbf{P}ooling (\textbf{PL}) module. 
$C_1$: Only pre-train the node embedding matrix $\boldsymbol{X}$ using VSubGAE in a \textbf{S}elf-\textbf{S}upervised (\textbf{SS}) manner. The position embedding matrix $\boldsymbol{P}$ is initialized randomly.
$C_2$: Only train the \textbf{P}osition \textbf{E}ncoding (\textbf{PE}). 
$C_3$: Initialize node embedding matrix $\boldsymbol{X}$ and position embedding matrix $\boldsymbol{P}$ randomly, and train PADEL by \textbf{C}ontrastive \textbf{L}earning (\textbf{CL}).
The subgraph generator VSubGAE 
is randomly initialized during training.
$C_4$: Remove \textbf{CL}, but \textbf{SS} and \textbf{PE} are retained.
$C_5$: Remove \textbf{PE}, but \textbf{SS} and \textbf{CL} are retained. The outputs of VSubGAE are fed into \textbf{CL} without position embedding.
$C_6$: Remove \textbf{SS}, use the position embedding matrix $\boldsymbol{P}$ as input features.
$C_7$: Use all components.

As shown in Table \ref{tab:Ablation}, each component of PADEL has a positive impact on results. The combination of all components brings the best results, which is much better than using any component alone.
Using two components is better than using a single component, except for $C_5$. It indicates that PE contributes the most to the model, as can be seen from the results of $C_2$, $C_4$, and $C_6$.
We observe that $C_1$ is worse than $C_0$ on the EM-USER dataset.  
We conduct an in-depth investigation and find that the subgraphs in EM-USER dataset are much larger than those in the other two datasets, so VSubGAE will truncate some of the nodes in the input, resulting in a high VSubGAE reconstruction loss. 


\begin{table}[]\small
\caption{We implement 8 variants of PADEL ($C_0$-$C_7$). The abbreviations in table: Self-Supervised (SS) learning, Position Encoding (PE), Contrastive Learning (CL), Pooling (PL).}\label{tab:Ablation}
\vspace{-0.4cm}
\centering
\setlength{\tabcolsep}{1mm}{
\begin{tabular}{lccccccc}
\bottomrule
SS & PE & CL & PL &   & \makecell{HPO-METAB \\ (10\%)} & \makecell{EM-USER \\(10\%)} & \makecell{HPO-NEURO \\(10\%)} \\ \midrule 
  &   &   & \multicolumn{1}{c|}{Y} & \multicolumn{1}{c|}{$C_0$} & 31.28±4.11 & 53.12±8.79  &43.67±2.33   \\
Y &   &   & \multicolumn{1}{c|}{Y} & \multicolumn{1}{c|}{$C_1$} & 35.80±2.80 & 52.00±9.75  &45.07±1.09   \\
  & Y &   & \multicolumn{1}{c|}{Y} & \multicolumn{1}{c|}{$C_2$} & 39.10±3.16 & 62.67±8.36  &49.82±1.05   \\
  &   & Y & \multicolumn{1}{c|}{Y} & \multicolumn{1}{c|}{$C_3$} & 33.49±2.88 & 59.11±3.61  &47.99±1.28   \\
Y & Y &   & \multicolumn{1}{c|}{Y} & \multicolumn{1}{c|}{$C_4$} & 40.21±3.60 & 63.78±6.21  &50.43±1.61  \\
Y &  & Y  & \multicolumn{1}{c|}{Y} & \multicolumn{1}{c|}{$C_5$} & 38.55±2.13 & 56.45±5.37  &47.15±1.41 \\
  & Y & Y & \multicolumn{1}{c|}{Y} & \multicolumn{1}{c|}{$C_6$} & 42.47±1.88 & 65.33±5.46  &52.04±1.37   \\
Y & Y & Y & \multicolumn{1}{c|}{Y} & \multicolumn{1}{c|}{$C_7$} & \textbf{44.72 ±1.34} & \textbf{68.45±5.51}  &\textbf{53.33±0.75}   \\ \bottomrule
\end{tabular}}
\end{table}

\begin{figure}[]
	\centering
	\subfigure[Our Method]{
		\begin{minipage}[t]{0.48\linewidth}
			\centering
			\includegraphics[width=\linewidth]{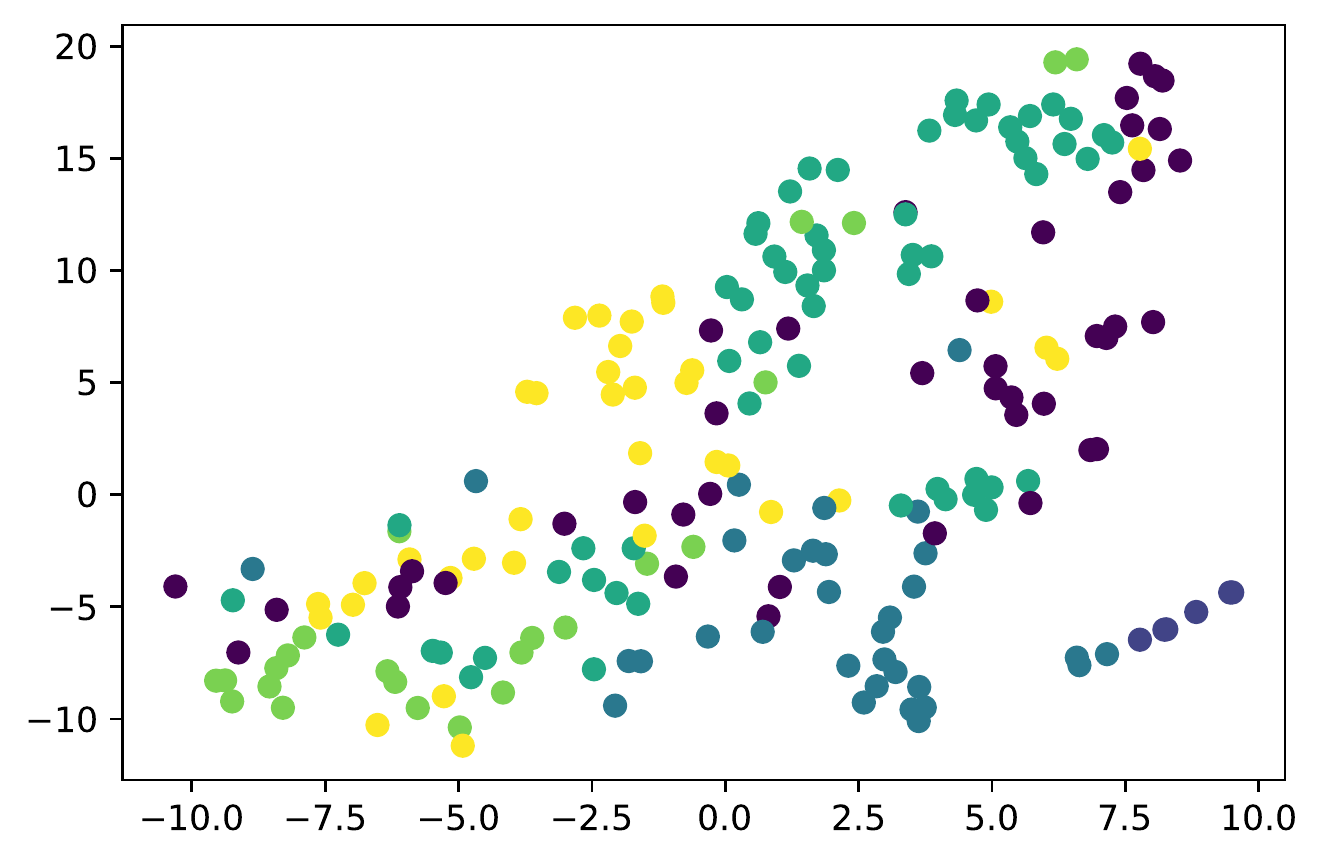}
		\end{minipage}%
		\label{fig:case_study_a}
	}%
	\subfigure[\textit{w.o.} PE \& Contrastive Learning]{
		\begin{minipage}[t]{0.48\linewidth}
			\centering
			\includegraphics[width=\linewidth]{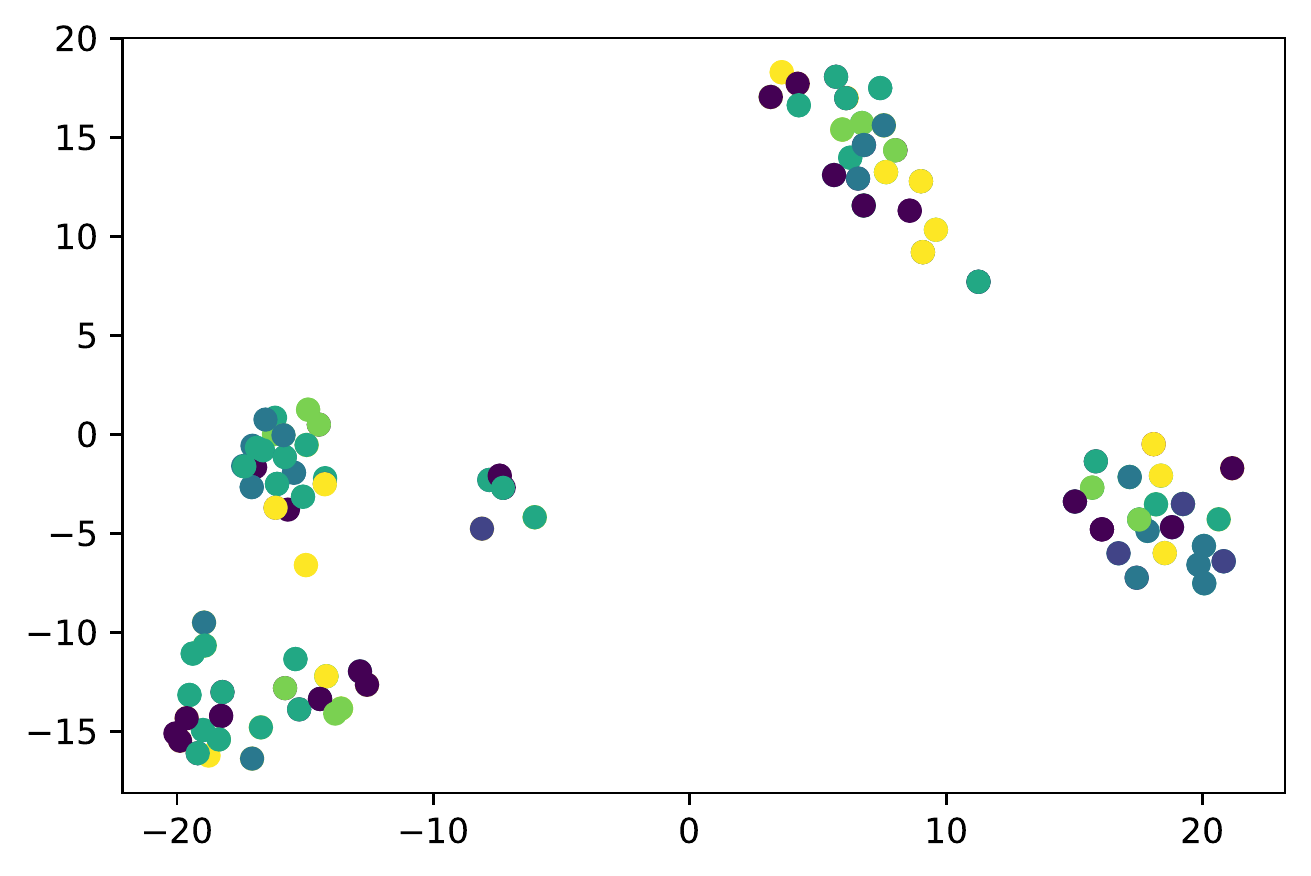}
		\end{minipage}%
		\label{fig:case_study_b}
	}%
	\centering
	\vspace{-0.4cm}
	\caption{Comparison of subgraph embeddings. Points in (a) are projections learned by PADEL. Points in (b) are learned without position encoding and contrastive learning.}
	\label{fig:case_study}
	\vspace{-0.4cm}
\end{figure}
\begin{figure}[]
	\centering
	\subfigure[3D position embedding.]{
		\begin{minipage}[t]{0.48\linewidth}
			\centering
			\includegraphics[width=\linewidth]{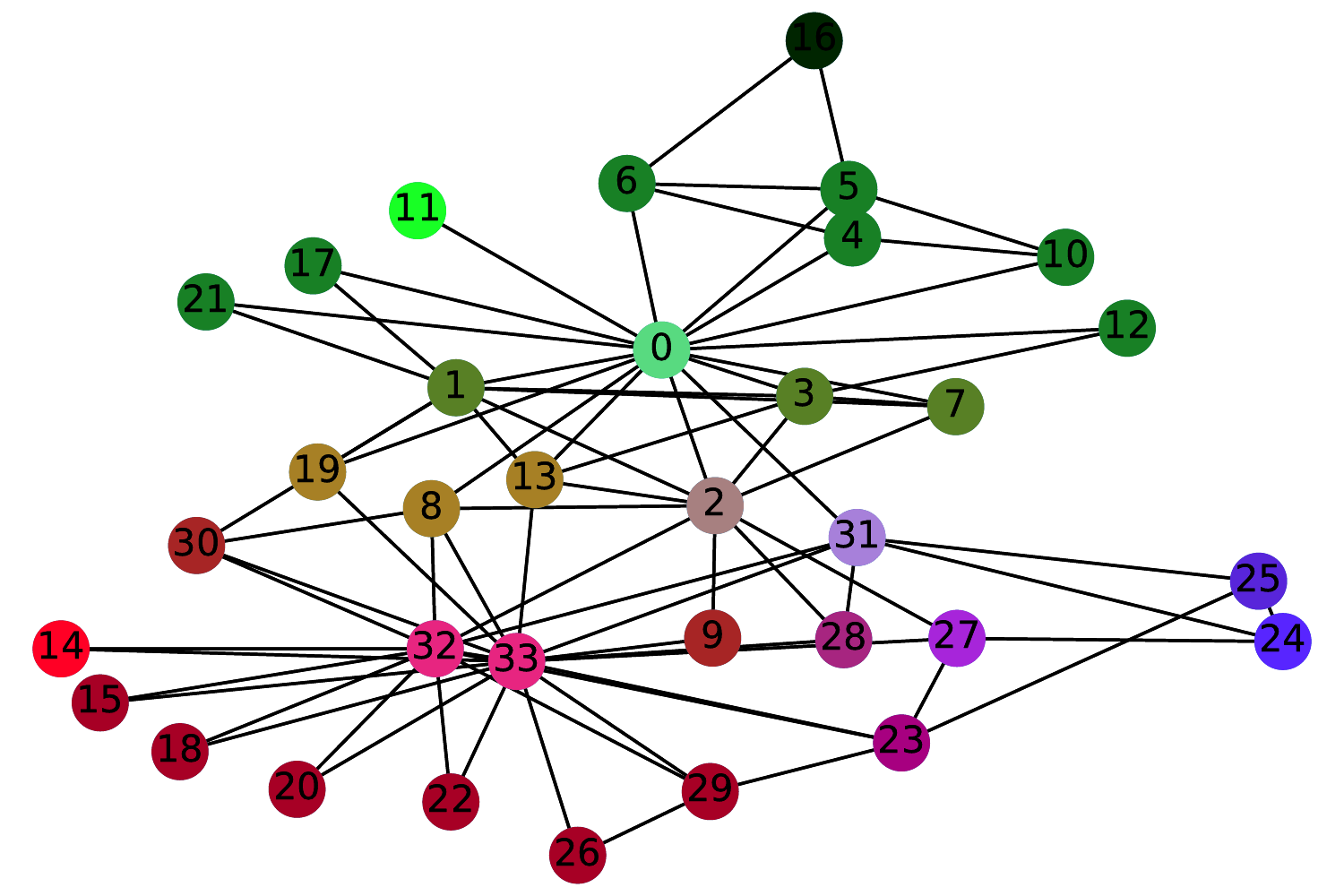}
		\end{minipage}%
		\label{fig:PE_RGB}
	}%
	\subfigure[2D position embedding.]{
		\begin{minipage}[t]{0.48\linewidth}
			\centering
			\includegraphics[width=\linewidth]{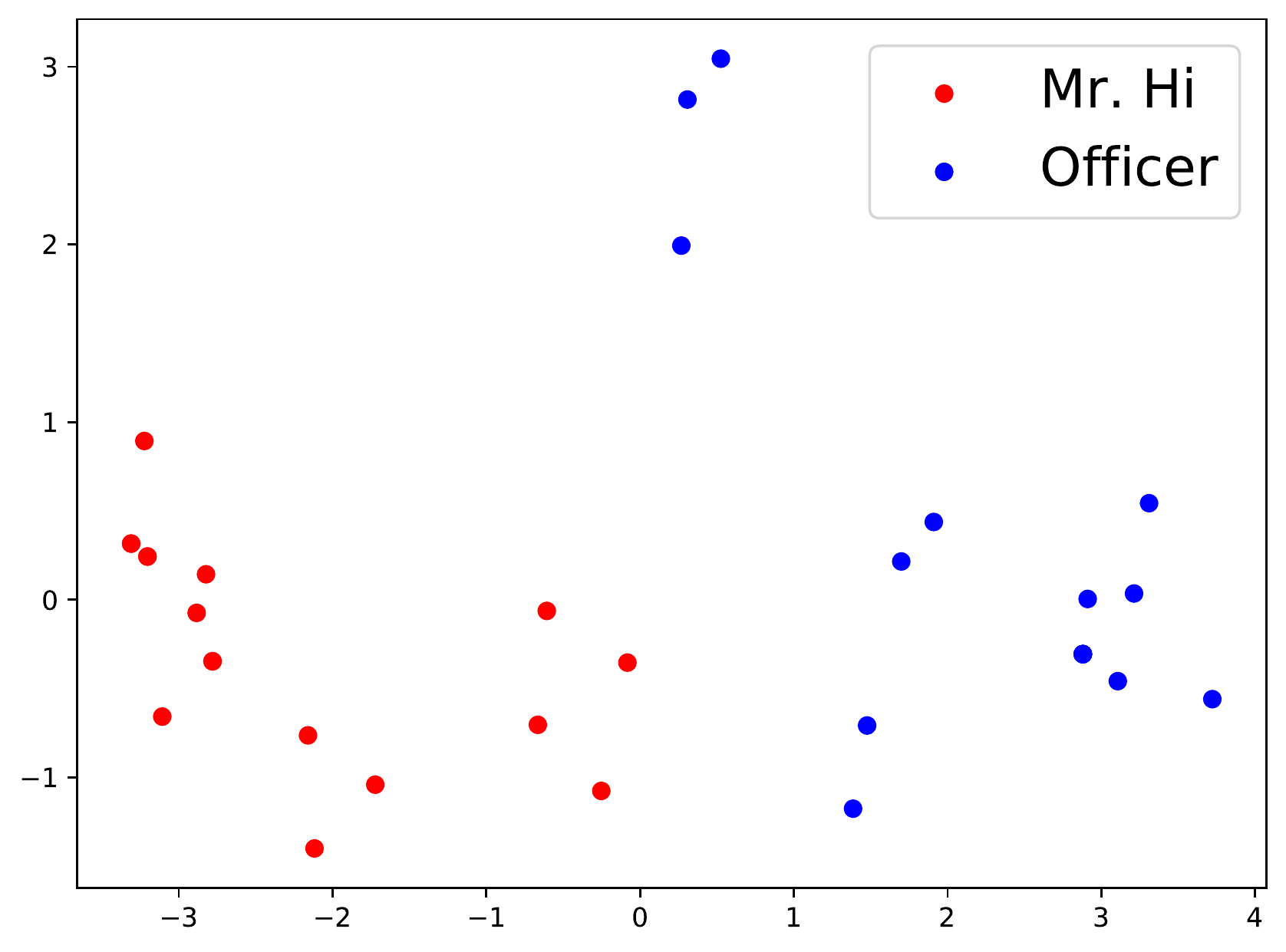}
		\end{minipage}
		\label{fig:PE_label}
	}%
	\centering
	
	\caption{Case study of cosine position encoding $\boldsymbol{P}$. (a) Position encoding as RGB value to color each node. Adjacent nodes have similar colors. (b) The 2-dimensional position embedding for each node. Node color illustrates its label.}  
	\label{fig:PE}
\end{figure}

\subsection{Case Study}
\label{case_study}
This section attempts to understand how PADEL facilitates subgraph representation learning. 
We select subgraphs in the HPO-METAB dataset, obtain their feature vectors after subgraph pooling, and visualize them via t-SNE \cite{tSNE} projection. 
Figure \ref{fig:case_study_a} illustrates the results learned by PADEL, and Figure \ref{fig:case_study_b} illustrates the results using simple end-to-end training without position encoding and contrastive learning. 
Each point represents a subgraph, and different colors represent different labels.
We can find that subgraphs with the same label are more likely to cluster together in Figure \ref{fig:case_study_a}, whereas subgraphs are entangled and hard to distinguish in Figure \ref{fig:case_study_b}.
It indicates that position encoding and contrastive learning help the model learning subgraph representation.

We provide a toy example on the Zachary's karate club network \cite{karate} to illustrate the expressiveness of our PE. We apply 3-dimensional cosine position embedding for each node in the graph. We categorize each node's position encoding as a Red-Green-Blue value and color each node to visualize them.  
Figure \ref{fig:PE_RGB} illustrates the results. We can observe that nodes close to each other have similar colors.
Further, we reduce the dimension to 2 and plot the scatter of $\boldsymbol{P}$. The results are shown in Figure \ref{fig:PE_label}. The position encoding makes nodes linearly separable \textbf{without supervised signal}. 

\section{Conclusion}
Learning subgraph neural networks given ``small'' data is a challenging task. This paper proposes a novel position-aware data-efficient subgraph neural networks called PADEL to solve the problem. We develop a novel position encoding method that is simple and powerful, design a generative subgraph augmentation method utilizing a variational subgraph autoencoder with random 1-hop subgraph diffusion, and develop exploratory and exploitable views for subgraph contrastive learning by a structure-aware pooling architecture. The self-supervised paradigm in PADEL helps to alleviate the bias problem in the subgraph prediction problem. Experiments on three real-world datasets show that our PADEL method surpasses the SOTA method by a large margin. 

\begin{acks}
This paper is supported by Huawei Technologies (Grant no. TC20201012001), NSFC (No. 62176155), Shanghai Municipal Science and Technology Major Project, China, under grant no. 2021SHZDZX0102.
\end{acks}

\clearpage
\bibliographystyle{ACM-Reference-Format}
\bibliography{acmart}


\begin{thebibliography}{56}


\ifx \showCODEN    \undefined \def \showCODEN     #1{\unskip}     \fi
\ifx \showDOI      \undefined \def \showDOI       #1{#1}\fi
\ifx \showISBNx    \undefined \def \showISBNx     #1{\unskip}     \fi
\ifx \showISBNxiii \undefined \def \showISBNxiii  #1{\unskip}     \fi
\ifx \showISSN     \undefined \def \showISSN      #1{\unskip}     \fi
\ifx \showLCCN     \undefined \def \showLCCN      #1{\unskip}     \fi
\ifx \shownote     \undefined \def \shownote      #1{#1}          \fi
\ifx \showarticletitle \undefined \def \showarticletitle #1{#1}   \fi
\ifx \showURL      \undefined \def \showURL       {\relax}        \fi
\providecommand\bibfield[2]{#2}
\providecommand\bibinfo[2]{#2}
\providecommand\natexlab[1]{#1}
\providecommand\showeprint[2][]{arXiv:#2}

\bibitem[Alsentzer et~al\mbox{.}(2020)]%
        {SubGNN}
\bibfield{author}{\bibinfo{person}{Emily Alsentzer}, \bibinfo{person}{Samuel~G.
  Finlayson}, \bibinfo{person}{Michelle~M. Li}, {and} \bibinfo{person}{Marinka
  Zitnik}.} \bibinfo{year}{2020}\natexlab{}.
\newblock \showarticletitle{Subgraph Neural Networks}. In
  \bibinfo{booktitle}{\emph{NeurIPS}}.
\newblock


\bibitem[Austin and Dawkins(2017)]%
        {Austin2017MedicalRN}
\bibfield{author}{\bibinfo{person}{Chris~P. Austin} {and} \bibinfo{person}{Hugh
  J.~S. Dawkins}.} \bibinfo{year}{2017}\natexlab{}.
\newblock \showarticletitle{Medical research: Next decade's goals for rare
  diseases}.
\newblock \bibinfo{journal}{\emph{NATURE}}  \bibinfo{volume}{548}
  (\bibinfo{year}{2017}), \bibinfo{pages}{158--158}.
\newblock


\bibitem[Bourgain(1985)]%
        {bourgain1985lipschitz}
\bibfield{author}{\bibinfo{person}{Jean Bourgain}.}
  \bibinfo{year}{1985}\natexlab{}.
\newblock \showarticletitle{On Lipschitz embedding of finite metric spaces in
  Hilbert space}.
\newblock \bibinfo{journal}{\emph{ISR J MATH}} \bibinfo{volume}{52},
  \bibinfo{number}{1-2} (\bibinfo{year}{1985}), \bibinfo{pages}{46--52}.
\newblock


\bibitem[Bradley(2020)]%
        {bradley2020statistical}
\bibfield{author}{\bibinfo{person}{Conor~A Bradley}.}
  \bibinfo{year}{2020}\natexlab{}.
\newblock \showarticletitle{A statistical framework for rare disease
  diagnosis}.
\newblock \bibinfo{journal}{\emph{NAT REV GENET}} \bibinfo{volume}{21},
  \bibinfo{number}{1} (\bibinfo{year}{2020}), \bibinfo{pages}{2--3}.
\newblock


\bibitem[Chang et~al\mbox{.}(2022)]%
        {Megnn}
\bibfield{author}{\bibinfo{person}{Yaomin Chang}, \bibinfo{person}{Chuan Chen},
  \bibinfo{person}{Weibo Hu}, \bibinfo{person}{Zibin Zheng},
  \bibinfo{person}{Xiaocong Zhou}, {and} \bibinfo{person}{Shouzhi Chen}.}
  \bibinfo{year}{2022}\natexlab{}.
\newblock \showarticletitle{Megnn: Meta-path extracted graph neural network for
  heterogeneous graph representation learning}.
\newblock \bibinfo{journal}{\emph{KNOWL BASED SYST}}  \bibinfo{volume}{235}
  (\bibinfo{year}{2022}), \bibinfo{pages}{107611}.
\newblock


\bibitem[Chen et~al\mbox{.}(2020)]%
        {bias}
\bibfield{author}{\bibinfo{person}{Jiawei Chen}, \bibinfo{person}{Hande Dong},
  \bibinfo{person}{Xiang Wang}, \bibinfo{person}{Fuli Feng},
  \bibinfo{person}{Meng Wang}, {and} \bibinfo{person}{Xiangnan He}.}
  \bibinfo{year}{2020}\natexlab{}.
\newblock \showarticletitle{Bias and Debias in Recommender System: {A} Survey
  and Future Directions}.
\newblock \bibinfo{journal}{\emph{arXiv preprint}}
  \bibinfo{volume}{arXiv:2010.03240} (\bibinfo{year}{2020}).
\newblock


\bibitem[Dijkstra(1959)]%
        {Dijkstra}
\bibfield{author}{\bibinfo{person}{Edsger~W. Dijkstra}.}
  \bibinfo{year}{1959}\natexlab{}.
\newblock \showarticletitle{A note on two problems in connexion with graphs}.
\newblock \bibinfo{journal}{\emph{NUMER MATH}}  \bibinfo{volume}{1}
  (\bibinfo{year}{1959}), \bibinfo{pages}{269--271}.
\newblock


\bibitem[et~al.(2019a)]%
        {Pytorch}
\bibfield{author}{\bibinfo{person}{Adam~Paszke et al.}}
  \bibinfo{year}{2019}\natexlab{a}.
\newblock \showarticletitle{PyTorch: An Imperative Style, High-Performance Deep
  Learning Library}. In \bibinfo{booktitle}{\emph{NeurIPS}}.
  \bibinfo{pages}{8024--8035}.
\newblock


\bibitem[et~al.(2019b)]%
        {Khler2019ExpansionOT}
\bibfield{author}{\bibinfo{person}{Sebastian~K{\"o}hler et al.}}
  \bibinfo{year}{2019}\natexlab{b}.
\newblock \showarticletitle{Expansion of the Human Phenotype Ontology (HPO)
  knowledge base and resources}.
\newblock \bibinfo{journal}{\emph{NUCLEIC ACIDS RES}}  \bibinfo{volume}{47}
  (\bibinfo{year}{2019}), \bibinfo{pages}{D1018 -- D1027}.
\newblock


\bibitem[Girvan and Newman(2002)]%
        {karate}
\bibfield{author}{\bibinfo{person}{Michelle Girvan} {and}
  \bibinfo{person}{Mark~EJ Newman}.} \bibinfo{year}{2002}\natexlab{}.
\newblock \showarticletitle{Community structure in social and biological
  networks}.
\newblock \bibinfo{journal}{\emph{P NATL ACAD SCI}} \bibinfo{volume}{99},
  \bibinfo{number}{12} (\bibinfo{year}{2002}), \bibinfo{pages}{7821--7826}.
\newblock


\bibitem[Graves et~al\mbox{.}(2005)]%
        {bilstm}
\bibfield{author}{\bibinfo{person}{Alex Graves}, \bibinfo{person}{Santiago
  Fern{\'{a}}ndez}, {and} \bibinfo{person}{J{\"{u}}rgen Schmidhuber}.}
  \bibinfo{year}{2005}\natexlab{}.
\newblock \showarticletitle{Bidirectional {LSTM} Networks for Improved Phoneme
  Classification and Recognition}. In \bibinfo{booktitle}{\emph{{ICANN}
  {(2)}}}, Vol.~\bibinfo{volume}{3697}. \bibinfo{pages}{799--804}.
\newblock


\bibitem[Hagberg et~al\mbox{.}(2008)]%
        {NetworkX}
\bibfield{author}{\bibinfo{person}{Aric~A. Hagberg}, \bibinfo{person}{Daniel~A.
  Schult}, {and} \bibinfo{person}{Pieter~J. Swart}.}
  \bibinfo{year}{2008}\natexlab{}.
\newblock \showarticletitle{Exploring Network Structure, Dynamics, and Function
  using NetworkX}. In \bibinfo{booktitle}{\emph{SciPy2008}}.
  \bibinfo{address}{Pasadena, CA USA}, \bibinfo{pages}{11 -- 15}.
\newblock


\bibitem[Hamilton et~al\mbox{.}(2017)]%
        {GraphRL}
\bibfield{author}{\bibinfo{person}{William~L. Hamilton}, \bibinfo{person}{Rex
  Ying}, {and} \bibinfo{person}{Jure Leskovec}.}
  \bibinfo{year}{2017}\natexlab{}.
\newblock \showarticletitle{Representation Learning on Graphs: Methods and
  Applications}.
\newblock \bibinfo{journal}{\emph{{IEEE} DATA ENG BULL}} \bibinfo{volume}{40},
  \bibinfo{number}{3} (\bibinfo{year}{2017}), \bibinfo{pages}{52--74}.
\newblock


\bibitem[Hartley et~al\mbox{.}(2020)]%
        {Hartley2020NewDA}
\bibfield{author}{\bibinfo{person}{Taila Hartley}, \bibinfo{person}{Gabrielle
  Lemire}, \bibinfo{person}{Kristin~D. Kernohan}, \bibinfo{person}{Heather~E
  Howley}, \bibinfo{person}{David~R. Adams}, {and} \bibinfo{person}{Kym~M.
  Boycott}.} \bibinfo{year}{2020}\natexlab{}.
\newblock \showarticletitle{New Diagnostic Approaches for Undiagnosed Rare
  Genetic Diseases.}
\newblock \bibinfo{journal}{\emph{ANNU REV GENOM HUM G}}
  (\bibinfo{year}{2020}).
\newblock


\bibitem[Hassani and Ahmadi(2020)]%
        {MVGRL}
\bibfield{author}{\bibinfo{person}{Kaveh Hassani} {and} \bibinfo{person}{Amir
  Hosein~Khas Ahmadi}.} \bibinfo{year}{2020}\natexlab{}.
\newblock \showarticletitle{Contrastive Multi-View Representation Learning on
  Graphs}. In \bibinfo{booktitle}{\emph{ICML}}, Vol.~\bibinfo{volume}{119}.
  \bibinfo{pages}{4116--4126}.
\newblock


\bibitem[Higgins et~al\mbox{.}(2017)]%
        {betaVAE}
\bibfield{author}{\bibinfo{person}{Irina Higgins}, \bibinfo{person}{Lo{\"{\i}}c
  Matthey}, \bibinfo{person}{Arka Pal}, \bibinfo{person}{Christopher Burgess},
  \bibinfo{person}{Xavier Glorot}, \bibinfo{person}{Matthew Botvinick},
  \bibinfo{person}{Shakir Mohamed}, {and} \bibinfo{person}{Alexander
  Lerchner}.} \bibinfo{year}{2017}\natexlab{}.
\newblock \showarticletitle{beta-VAE: Learning Basic Visual Concepts with a
  Constrained Variational Framework}. In \bibinfo{booktitle}{\emph{{ICLR}
  (Poster)}}.
\newblock


\bibitem[Hwang et~al\mbox{.}(2020)]%
        {SSAL}
\bibfield{author}{\bibinfo{person}{Dasol Hwang}, \bibinfo{person}{Jinyoung
  Park}, \bibinfo{person}{Sunyoung Kwon}, \bibinfo{person}{Kyung{-}Min Kim},
  \bibinfo{person}{Jung{-}Woo Ha}, {and} \bibinfo{person}{Hyunwoo~J. Kim}.}
  \bibinfo{year}{2020}\natexlab{}.
\newblock \showarticletitle{Self-supervised Auxiliary Learning with Meta-paths
  for Heterogeneous Graphs}. In \bibinfo{booktitle}{\emph{NeurIPS}}.
\newblock


\bibitem[Jiang et~al\mbox{.}(2021)]%
        {MetaP}
\bibfield{author}{\bibinfo{person}{Zhiyi Jiang}, \bibinfo{person}{Jianliang
  Gao}, {and} \bibinfo{person}{Xinqi Lv}.} \bibinfo{year}{2021}\natexlab{}.
\newblock \showarticletitle{MetaP: Meta Pattern Learning for One-Shot Knowledge
  Graph Completion}. In \bibinfo{booktitle}{\emph{SIGIR}}.
  \bibinfo{pages}{2232--2236}.
\newblock


\bibitem[Joshi and Urbani(2020)]%
        {Sub-kg-1}
\bibfield{author}{\bibinfo{person}{Unmesh Joshi} {and} \bibinfo{person}{Jacopo
  Urbani}.} \bibinfo{year}{2020}\natexlab{}.
\newblock \showarticletitle{Searching for Embeddings in a Haystack: Link
  Prediction on Knowledge Graphs with Subgraph Pruning}. In
  \bibinfo{booktitle}{\emph{WWW}}. \bibinfo{pages}{2817--2823}.
\newblock


\bibitem[Kim and Price(2011)]%
        {Kim2011GeneticCN}
\bibfield{author}{\bibinfo{person}{Pan-Jun Kim} {and}
  \bibinfo{person}{Nathan~D. Price}.} \bibinfo{year}{2011}\natexlab{}.
\newblock \showarticletitle{Genetic Co-Occurrence Network across Sequenced
  Microbes}.
\newblock \bibinfo{journal}{\emph{PLOS COMPUT BIOL}}  \bibinfo{volume}{7}
  (\bibinfo{year}{2011}).
\newblock


\bibitem[Kipf and Welling(2016)]%
        {VGAE}
\bibfield{author}{\bibinfo{person}{Thomas~N. Kipf} {and} \bibinfo{person}{Max
  Welling}.} \bibinfo{year}{2016}\natexlab{}.
\newblock \showarticletitle{Variational Graph Auto-Encoders}.
\newblock \bibinfo{journal}{\emph{arXiv preprint}}
  \bibinfo{volume}{arXiv:1611.07308} (\bibinfo{year}{2016}).
\newblock


\bibitem[Kipf and Welling(2017)]%
        {GCN}
\bibfield{author}{\bibinfo{person}{Thomas~N. Kipf} {and} \bibinfo{person}{Max
  Welling}.} \bibinfo{year}{2017}\natexlab{}.
\newblock \showarticletitle{Semi-Supervised Classification with Graph
  Convolutional Networks}. In \bibinfo{booktitle}{\emph{{ICLR} (Poster)}}.
\newblock


\bibitem[Lai et~al\mbox{.}(2021)]%
        {ARCLink}
\bibfield{author}{\bibinfo{person}{Darong Lai}, \bibinfo{person}{Zheyi Liu},
  \bibinfo{person}{Junyao Huang}, \bibinfo{person}{Zhihong Chong},
  \bibinfo{person}{Weiwei Wu}, {and} \bibinfo{person}{Christine Nardini}.}
  \bibinfo{year}{2021}\natexlab{}.
\newblock \showarticletitle{Attention Based Subgraph Classification for Link
  Prediction by Network Re-weighting}. In \bibinfo{booktitle}{\emph{CIKM}}.
  \bibinfo{pages}{3171--3175}.
\newblock


\bibitem[Li et~al\mbox{.}(2020)]%
        {DEGNN}
\bibfield{author}{\bibinfo{person}{Pan Li}, \bibinfo{person}{Yanbang Wang},
  \bibinfo{person}{Hongwei Wang}, {and} \bibinfo{person}{Jure Leskovec}.}
  \bibinfo{year}{2020}\natexlab{}.
\newblock \showarticletitle{Distance Encoding: Design Provably More Powerful
  Neural Networks for Graph Representation Learning}. In
  \bibinfo{booktitle}{\emph{NeurIPS}}.
\newblock


\bibitem[Liu et~al\mbox{.}(2020)]%
        {SHFF}
\bibfield{author}{\bibinfo{person}{Zheyi Liu}, \bibinfo{person}{Darong Lai},
  \bibinfo{person}{Chuanyou Li}, {and} \bibinfo{person}{Meng Wang}.}
  \bibinfo{year}{2020}\natexlab{}.
\newblock \showarticletitle{Feature Fusion Based Subgraph Classification for
  Link Prediction}. In \bibinfo{booktitle}{\emph{CIKM}}.
  \bibinfo{pages}{985--994}.
\newblock


\bibitem[Loshchilov and Hutter(2019)]%
        {AdamW}
\bibfield{author}{\bibinfo{person}{Ilya Loshchilov} {and}
  \bibinfo{person}{Frank Hutter}.} \bibinfo{year}{2019}\natexlab{}.
\newblock \showarticletitle{Decoupled Weight Decay Regularization}. In
  \bibinfo{booktitle}{\emph{{ICLR} (Poster)}}.
\newblock


\bibitem[Marinka~Zitnik and Leskovec(2018)]%
        {zitnik2018biosnap}
\bibfield{author}{\bibinfo{person}{Sagar~Maheshwari Marinka~Zitnik,
  Rok~Sosi\v{c}} {and} \bibinfo{person}{Jure Leskovec}.}
  \bibinfo{year}{2018}\natexlab{}.
\newblock \bibinfo{title}{{BioSNAP Datasets}: {Stanford} Biomedical Network
  Dataset Collection}.
\newblock \bibinfo{howpublished}{\url{http://snap.stanford.edu/biodata}}.
\newblock


\bibitem[Meng et~al\mbox{.}(2018)]%
        {SPNN}
\bibfield{author}{\bibinfo{person}{Changping Meng}, \bibinfo{person}{S.~Chandra
  Mouli}, \bibinfo{person}{Bruno Ribeiro}, {and} \bibinfo{person}{Jennifer
  Neville}.} \bibinfo{year}{2018}\natexlab{}.
\newblock \showarticletitle{Subgraph Pattern Neural Networks for High-Order
  Graph Evolution Prediction}. In \bibinfo{booktitle}{\emph{{AAAI}}}.
  \bibinfo{pages}{3778--3787}.
\newblock


\bibitem[Mesquita et~al\mbox{.}(2020)]%
        {RethinkPooling}
\bibfield{author}{\bibinfo{person}{Diego P.~P. Mesquita},
  \bibinfo{person}{Amauri H.~Souza Jr.}, {and} \bibinfo{person}{Samuel Kaski}.}
  \bibinfo{year}{2020}\natexlab{}.
\newblock \showarticletitle{Rethinking pooling in graph neural networks}. In
  \bibinfo{booktitle}{\emph{NeurIPS}}.
\newblock


\bibitem[Ni et~al\mbox{.}(2019)]%
        {Endomondo}
\bibfield{author}{\bibinfo{person}{Jianmo Ni}, \bibinfo{person}{Larry
  Muhlstein}, {and} \bibinfo{person}{Julian~J. McAuley}.}
  \bibinfo{year}{2019}\natexlab{}.
\newblock \showarticletitle{Modeling Heart Rate and Activity Data for
  Personalized Fitness Recommendation}. In \bibinfo{booktitle}{\emph{{WWW}}}.
  \bibinfo{pages}{1343--1353}.
\newblock


\bibitem[Qiu et~al\mbox{.}(2020)]%
        {GCC}
\bibfield{author}{\bibinfo{person}{Jiezhong Qiu}, \bibinfo{person}{Qibin Chen},
  \bibinfo{person}{Yuxiao Dong}, \bibinfo{person}{Jing Zhang},
  \bibinfo{person}{Hongxia Yang}, \bibinfo{person}{Ming Ding},
  \bibinfo{person}{Kuansan Wang}, {and} \bibinfo{person}{Jie Tang}.}
  \bibinfo{year}{2020}\natexlab{}.
\newblock \showarticletitle{{GCC:} Graph Contrastive Coding for Graph Neural
  Network Pre-Training}. In \bibinfo{booktitle}{\emph{{KDD}}}.
  \bibinfo{pages}{1150--1160}.
\newblock


\bibitem[Rong et~al\mbox{.}(2020)]%
        {GROVER}
\bibfield{author}{\bibinfo{person}{Yu Rong}, \bibinfo{person}{Yatao Bian},
  \bibinfo{person}{Tingyang Xu}, \bibinfo{person}{Weiyang Xie},
  \bibinfo{person}{Ying Wei}, \bibinfo{person}{Wenbing Huang}, {and}
  \bibinfo{person}{Junzhou Huang}.} \bibinfo{year}{2020}\natexlab{}.
\newblock \showarticletitle{Self-Supervised Graph Transformer on Large-Scale
  Molecular Data}. In \bibinfo{booktitle}{\emph{NeurIPS}}.
\newblock


\bibitem[Sun et~al\mbox{.}(2020)]%
        {InfoGraph}
\bibfield{author}{\bibinfo{person}{Fan{-}Yun Sun}, \bibinfo{person}{Jordan
  Hoffmann}, \bibinfo{person}{Vikas Verma}, {and} \bibinfo{person}{Jian Tang}.}
  \bibinfo{year}{2020}\natexlab{}.
\newblock \showarticletitle{InfoGraph: Unsupervised and Semi-supervised
  Graph-Level Representation Learning via Mutual Information Maximization}. In
  \bibinfo{booktitle}{\emph{{ICLR}}}.
\newblock


\bibitem[Sun et~al\mbox{.}(2021)]%
        {SUGAR}
\bibfield{author}{\bibinfo{person}{Qingyun Sun}, \bibinfo{person}{Jianxin Li},
  \bibinfo{person}{Hao Peng}, \bibinfo{person}{Jia Wu},
  \bibinfo{person}{Yuanxing Ning}, \bibinfo{person}{Philip~S. Yu}, {and}
  \bibinfo{person}{Lifang He}.} \bibinfo{year}{2021}\natexlab{}.
\newblock \showarticletitle{{SUGAR:} Subgraph Neural Network with Reinforcement
  Pooling and Self-Supervised Mutual Information Mechanism}. In
  \bibinfo{booktitle}{\emph{{WWW}}}. \bibinfo{publisher}{{ACM} / {IW3C2}},
  \bibinfo{pages}{2081--2091}.
\newblock


\bibitem[Teru et~al\mbox{.}(2020)]%
        {GraIL}
\bibfield{author}{\bibinfo{person}{Komal~K. Teru}, \bibinfo{person}{Etienne
  Denis}, {and} \bibinfo{person}{Will Hamilton}.}
  \bibinfo{year}{2020}\natexlab{}.
\newblock \showarticletitle{Inductive Relation Prediction by Subgraph
  Reasoning}. In \bibinfo{booktitle}{\emph{ICML}}, Vol.~\bibinfo{volume}{119}.
  \bibinfo{pages}{9448--9457}.
\newblock


\bibitem[Van~der Maaten and Hinton(2008)]%
        {tSNE}
\bibfield{author}{\bibinfo{person}{Laurens Van~der Maaten} {and}
  \bibinfo{person}{Geoffrey Hinton}.} \bibinfo{year}{2008}\natexlab{}.
\newblock \showarticletitle{Visualizing data using t-SNE.}
\newblock \bibinfo{journal}{\emph{JMLR}} \bibinfo{volume}{9},
  \bibinfo{number}{11} (\bibinfo{year}{2008}).
\newblock


\bibitem[Vaswani et~al\mbox{.}(2017)]%
        {transformer}
\bibfield{author}{\bibinfo{person}{Ashish Vaswani}, \bibinfo{person}{Noam
  Shazeer}, \bibinfo{person}{Niki Parmar}, \bibinfo{person}{Jakob Uszkoreit},
  \bibinfo{person}{Llion Jones}, \bibinfo{person}{Aidan~N. Gomez},
  \bibinfo{person}{Lukasz Kaiser}, {and} \bibinfo{person}{Illia Polosukhin}.}
  \bibinfo{year}{2017}\natexlab{}.
\newblock \showarticletitle{Attention is All you Need}. In
  \bibinfo{booktitle}{\emph{{NIPS}}}. \bibinfo{pages}{5998--6008}.
\newblock


\bibitem[Velickovic et~al\mbox{.}(2018)]%
        {GAT}
\bibfield{author}{\bibinfo{person}{Petar Velickovic}, \bibinfo{person}{Guillem
  Cucurull}, \bibinfo{person}{Arantxa Casanova}, \bibinfo{person}{Adriana
  Romero}, \bibinfo{person}{Pietro Li{\`{o}}}, {and} \bibinfo{person}{Yoshua
  Bengio}.} \bibinfo{year}{2018}\natexlab{}.
\newblock \showarticletitle{Graph Attention Networks}. In
  \bibinfo{booktitle}{\emph{{ICLR} (Poster)}}.
\newblock


\bibitem[Velickovic et~al\mbox{.}(2019)]%
        {DGI}
\bibfield{author}{\bibinfo{person}{Petar Velickovic}, \bibinfo{person}{William
  Fedus}, \bibinfo{person}{William~L. Hamilton}, \bibinfo{person}{Pietro
  Li{\`{o}}}, \bibinfo{person}{Yoshua Bengio}, {and} \bibinfo{person}{R.~Devon
  Hjelm}.} \bibinfo{year}{2019}\natexlab{}.
\newblock \showarticletitle{Deep Graph Infomax}. In
  \bibinfo{booktitle}{\emph{{ICLR} (Poster)}}.
\newblock


\bibitem[Wang et~al\mbox{.}(2021a)]%
        {SLiCE}
\bibfield{author}{\bibinfo{person}{Ping Wang}, \bibinfo{person}{Khushbu
  Agarwal}, \bibinfo{person}{Colby Ham}, \bibinfo{person}{Sutanay Choudhury},
  {and} \bibinfo{person}{Chandan~K. Reddy}.} \bibinfo{year}{2021}\natexlab{a}.
\newblock \showarticletitle{Self-Supervised Learning of Contextual Embeddings
  for Link Prediction in Heterogeneous Networks}. In
  \bibinfo{booktitle}{\emph{{WWW}}}. \bibinfo{pages}{2946--2957}.
\newblock


\bibitem[Wang et~al\mbox{.}(2021b)]%
        {HeCo}
\bibfield{author}{\bibinfo{person}{Xiao Wang}, \bibinfo{person}{Nian Liu},
  \bibinfo{person}{Hui Han}, {and} \bibinfo{person}{Chuan Shi}.}
  \bibinfo{year}{2021}\natexlab{b}.
\newblock \showarticletitle{Self-supervised Heterogeneous Graph Neural Network
  with Co-contrastive Learning}. In \bibinfo{booktitle}{\emph{KDD}}.
  \bibinfo{pages}{1726--1736}.
\newblock


\bibitem[Wang and Zhang(2022)]%
        {glass}
\bibfield{author}{\bibinfo{person}{Xiyuan Wang} {and} \bibinfo{person}{Muhan
  Zhang}.} \bibinfo{year}{2022}\natexlab{}.
\newblock \showarticletitle{{GLASS}: {GNN} with Labeling Tricks for Subgraph
  Representation Learning}. In \bibinfo{booktitle}{\emph{International
  Conference on Learning Representations}}.
\newblock
\urldef\tempurl%
\url{https://openreview.net/forum?id=XLxhEjKNbXj}
\showURL{%
\tempurl}


\bibitem[Wold et~al\mbox{.}(1987)]%
        {PCA}
\bibfield{author}{\bibinfo{person}{Svante Wold}, \bibinfo{person}{Kim
  Esbensen}, {and} \bibinfo{person}{Paul Geladi}.}
  \bibinfo{year}{1987}\natexlab{}.
\newblock \showarticletitle{Principal component analysis}.
\newblock \bibinfo{journal}{\emph{CHEMOMETR INTELL LAB}} \bibinfo{volume}{2},
  \bibinfo{number}{1-3} (\bibinfo{year}{1987}), \bibinfo{pages}{37--52}.
\newblock


\bibitem[Xiao et~al\mbox{.}(2021)]%
        {DBLP:conf/iclr/Xiao0ED21}
\bibfield{author}{\bibinfo{person}{Tete Xiao}, \bibinfo{person}{Xiaolong Wang},
  \bibinfo{person}{Alexei~A. Efros}, {and} \bibinfo{person}{Trevor Darrell}.}
  \bibinfo{year}{2021}\natexlab{}.
\newblock \showarticletitle{What Should Not Be Contrastive in Contrastive
  Learning}. In \bibinfo{booktitle}{\emph{{ICLR}}}.
\newblock


\bibitem[Xie et~al\mbox{.}(2021)]%
        {SSL_review}
\bibfield{author}{\bibinfo{person}{Yaochen Xie}, \bibinfo{person}{Zhao Xu},
  \bibinfo{person}{Zhengyang Wang}, {and} \bibinfo{person}{Shuiwang Ji}.}
  \bibinfo{year}{2021}\natexlab{}.
\newblock \showarticletitle{Self-Supervised Learning of Graph Neural Networks:
  {A} Unified Review}.
\newblock \bibinfo{journal}{\emph{arXiv preprint}}
  \bibinfo{volume}{arXiv:2102.10757} (\bibinfo{year}{2021}).
\newblock


\bibitem[Xu et~al\mbox{.}(2019)]%
        {GIN}
\bibfield{author}{\bibinfo{person}{Keyulu Xu}, \bibinfo{person}{Weihua Hu},
  \bibinfo{person}{Jure Leskovec}, {and} \bibinfo{person}{Stefanie Jegelka}.}
  \bibinfo{year}{2019}\natexlab{}.
\newblock \showarticletitle{How Powerful are Graph Neural Networks?}. In
  \bibinfo{booktitle}{\emph{{ICLR}}}.
\newblock


\bibitem[Xu et~al\mbox{.}(2021)]%
        {GraphLoG}
\bibfield{author}{\bibinfo{person}{Minghao Xu}, \bibinfo{person}{Hang Wang},
  \bibinfo{person}{Bingbing Ni}, \bibinfo{person}{Hongyu Guo}, {and}
  \bibinfo{person}{Jian Tang}.} \bibinfo{year}{2021}\natexlab{}.
\newblock \showarticletitle{Self-supervised Graph-level Representation Learning
  with Local and Global Structure}. In \bibinfo{booktitle}{\emph{ICML}},
  Vol.~\bibinfo{volume}{139}. \bibinfo{pages}{11548--11558}.
\newblock


\bibitem[You et~al\mbox{.}(2019)]%
        {PGNN}
\bibfield{author}{\bibinfo{person}{Jiaxuan You}, \bibinfo{person}{Rex Ying},
  {and} \bibinfo{person}{Jure Leskovec}.} \bibinfo{year}{2019}\natexlab{}.
\newblock \showarticletitle{Position-aware Graph Neural Networks}. In
  \bibinfo{booktitle}{\emph{ICML}}, Vol.~\bibinfo{volume}{97}.
  \bibinfo{pages}{7134--7143}.
\newblock


\bibitem[You et~al\mbox{.}(2021)]%
        {JOAO}
\bibfield{author}{\bibinfo{person}{Yuning You}, \bibinfo{person}{Tianlong
  Chen}, \bibinfo{person}{Yang Shen}, {and} \bibinfo{person}{Zhangyang Wang}.}
  \bibinfo{year}{2021}\natexlab{}.
\newblock \showarticletitle{Graph Contrastive Learning Automated}. In
  \bibinfo{booktitle}{\emph{ICML}}, Vol.~\bibinfo{volume}{139}.
  \bibinfo{pages}{12121--12132}.
\newblock


\bibitem[You et~al\mbox{.}(2020)]%
        {GraphCL}
\bibfield{author}{\bibinfo{person}{Yuning You}, \bibinfo{person}{Tianlong
  Chen}, \bibinfo{person}{Yongduo Sui}, \bibinfo{person}{Ting Chen},
  \bibinfo{person}{Zhangyang Wang}, {and} \bibinfo{person}{Yang Shen}.}
  \bibinfo{year}{2020}\natexlab{}.
\newblock \showarticletitle{Graph Contrastive Learning with Augmentations}. In
  \bibinfo{booktitle}{\emph{NeurIPS}}.
\newblock


\bibitem[Yuan et~al\mbox{.}(2021)]%
        {SubgraphX}
\bibfield{author}{\bibinfo{person}{Hao Yuan}, \bibinfo{person}{Haiyang Yu},
  \bibinfo{person}{Jie Wang}, \bibinfo{person}{Kang Li}, {and}
  \bibinfo{person}{Shuiwang Ji}.} \bibinfo{year}{2021}\natexlab{}.
\newblock \showarticletitle{On Explainability of Graph Neural Networks via
  Subgraph Explorations}. In \bibinfo{booktitle}{\emph{ICML}},
  Vol.~\bibinfo{volume}{139}. \bibinfo{pages}{12241--12252}.
\newblock


\bibitem[Zeng and Xie(2021)]%
        {CCSL}
\bibfield{author}{\bibinfo{person}{Jiaqi Zeng} {and} \bibinfo{person}{Pengtao
  Xie}.} \bibinfo{year}{2021}\natexlab{}.
\newblock \showarticletitle{Contrastive Self-supervised Learning for Graph
  Classification}. In \bibinfo{booktitle}{\emph{AAAI}}.
  \bibinfo{pages}{10824--10832}.
\newblock


\bibitem[Zhang et~al\mbox{.}(2021)]%
        {Data-efficient-KDD21}
\bibfield{author}{\bibinfo{person}{Chuxu Zhang}, \bibinfo{person}{Jundong Li},
  {and} \bibinfo{person}{Meng Jiang}.} \bibinfo{year}{2021}\natexlab{}.
\newblock \showarticletitle{Data Efficient Learning on Graphs}. In
  \bibinfo{booktitle}{\emph{{KDD}}}. \bibinfo{pages}{4092--4093}.
\newblock


\bibitem[Zhao et~al\mbox{.}(2021)]%
        {DBLP:conf/aaai/0003LNW0S21}
\bibfield{author}{\bibinfo{person}{Tong Zhao}, \bibinfo{person}{Yozen Liu},
  \bibinfo{person}{Leonardo Neves}, \bibinfo{person}{Oliver~J. Woodford},
  \bibinfo{person}{Meng Jiang}, {and} \bibinfo{person}{Neil Shah}.}
  \bibinfo{year}{2021}\natexlab{}.
\newblock \showarticletitle{Data Augmentation for Graph Neural Networks}. In
  \bibinfo{booktitle}{\emph{{AAAI}}}. \bibinfo{pages}{11015--11023}.
\newblock


\bibitem[Zhou et~al\mbox{.}(2020)]%
        {GNN_review_JZ}
\bibfield{author}{\bibinfo{person}{Jie Zhou}, \bibinfo{person}{Ganqu Cui},
  \bibinfo{person}{Shengding Hu}, \bibinfo{person}{Zhengyan Zhang},
  \bibinfo{person}{Cheng Yang}, \bibinfo{person}{Zhiyuan Liu},
  \bibinfo{person}{Lifeng Wang}, \bibinfo{person}{Changcheng Li}, {and}
  \bibinfo{person}{Maosong Sun}.} \bibinfo{year}{2020}\natexlab{}.
\newblock \showarticletitle{Graph neural networks: {A} review of methods and
  applications}.
\newblock \bibinfo{journal}{\emph{{AI} Open}}  \bibinfo{volume}{1}
  (\bibinfo{year}{2020}), \bibinfo{pages}{57--81}.
\newblock


\bibitem[Zhu et~al\mbox{.}(2021)]%
        {GCA}
\bibfield{author}{\bibinfo{person}{Yanqiao Zhu}, \bibinfo{person}{Yichen Xu},
  \bibinfo{person}{Feng Yu}, \bibinfo{person}{Qiang Liu}, \bibinfo{person}{Shu
  Wu}, {and} \bibinfo{person}{Liang Wang}.} \bibinfo{year}{2021}\natexlab{}.
\newblock \showarticletitle{Graph Contrastive Learning with Adaptive
  Augmentation}. In \bibinfo{booktitle}{\emph{{WWW}}}.
  \bibinfo{pages}{2069--2080}.
\newblock


\end{thebibliography}

\end{sloppypar}

\end{document}